\documentclass[10pt,journal,compsoc]{IEEEtran}

\usepackage{tabularx}
\usepackage{array}

\usepackage[switch]{lineno}
\usepackage{booktabs}
\usepackage{pgfplots}
\usepackage{mathrsfs}
\usepackage{amsmath}
\pgfplotsset{width=10cm,compat=1.9}
\usepgfplotslibrary{external}
\usepackage{pgfplots}
\usepackage{amsfonts}
\pgfplotsset{compat=newest}
\usetikzlibrary{decorations.pathmorphing}
\def\infinity{\rotatebox{90}{8}}
\usepackage{float}
\usepackage{pbox}
\usepackage{footnote}
\usepackage{multirow}

\usepackage[noend]{algpseudocode}
\makeatletter

\usepackage[hang,flushmargin]{footmisc}
\usepackage{lipsum}
\makeatletter
\newcommand{\algorithmfootnote}[2][\footnotesize]{%
  \let\old@algocf@finish\@algocf@finish% Store algorithm finish macro
  \def\@algocf@finish{\old@algocf@finish% Update finish macro to insert "footnote"
    \leavevmode\rlap{\begin{minipage}{\linewidth}
    #1#2
    \end{minipage}}%
  }%
}

\usepackage[linesnumbered, ruled]{algorithm2e}
\SetKwRepeat{Do}{do}{while}

\ifCLASSOPTIONcompsoc
  \usepackage[nocompress]{cite}
\else
  \usepackage{cite}
\fi
\ifCLASSINFOpdf
\else
\fi
\begin{document}
%
% \linenumbers
% paper title
% Titles are generally capitalized except for words such as a, an, and, as,
% at, but, by, for, in, nor, of, on, or, the, to and up, which are usually
% not capitalized unless they are the first or last word of the title.
% Linebreaks \\ can be used within to get better formatting as desired.
% Do not put math or special symbols in the title.
\title{Generalizable Data-free Objective for Crafting Universal Adversarial Perturbations}
%
%
% author names and IEEE memberships
% note positions of commas and nonbreaking spaces ( ~ ) LaTeX will not break
% a structure at a ~ so this keeps an author's name from being broken across
% two lines.
% use \thanks{} to gain access to the first footnote area
% a separate \thanks must be used for each paragraph as LaTeX2e's \thanks
% was not built to handle multiple paragraphs
%
%
%\IEEEcompsocitemizethanks is a special \thanks that produces the bulleted
% lists the Computer Society journals use for "first footnote" author
% affiliations. Use \IEEEcompsocthanksitem which works much like \item
% for each affiliation group. When not in compsoc mode,
% \IEEEcompsocitemizethanks becomes like \thanks and
% \IEEEcompsocthanksitem becomes a line break with idention. This
% facilitates dual compilation, although admittedly the differences in the
% desired content of \author between the different types of papers makes a
% one-size-fits-all approach a daunting prospect. For instance, compsoc 
% journal papers have the author affiliations above the "Manuscript
% received ..."  text while in non-compsoc journals this is reversed. Sigh.

\author{Konda Reddy Mopuri*,
        Aditya Ganeshan*,
        R. Venkatesh Babu,~\IEEEmembership{Senior~Member,~IEEE}% <-this % stops a space
\IEEEcompsocitemizethanks{\IEEEcompsocthanksitem The authors are with the Video Analytics Lab, Department of Computational and Data Sciences, Indian Institute of Science, Bangalore, India, 560012.\protect\\
% note need leading \protect in front of \\ to get a newline within \thanks as
% \\ is fragile and will error, could use \hfil\break instead.
E-mail: kondamopuri@iisc.ac.in, adityaganeshan@gmail.com and venky@iisc.ac.in
%\IEEEcompsocthanksitem J. Doe and J. Doe are with Anonymous University.
}% <-this % stops an unwanted space
%\thanks{Manuscript received April 19, 2005; revised August 26, 2015.}
}

\IEEEtitleabstractindextext{%
\begin{abstract}
Machine learning models are susceptible to adversarial perturbations: small changes to input that can cause large changes in output. It is also demonstrated that there exist input-agnostic perturbations, called universal adversarial perturbations, which can change the inference of target model on most of the data samples. However, existing methods to craft universal perturbations are (i) task specific, (ii) require samples from the training data distribution, and (iii) perform complex optimizations. Additionally, because of the data dependence, fooling ability of the crafted perturbations is proportional to the available training data. In this paper, we present a novel, generalizable and data-free approaches for crafting universal adversarial perturbations. Independent of the underlying task, our objective achieves fooling via corrupting the extracted features at multiple layers. Therefore, the proposed objective is generalizable to craft image-agnostic perturbations across multiple vision tasks such as object recognition, semantic segmentation, and depth estimation. In the practical setting of black-box attack scenario (when the attacker does not have access to the target model and it's training data), we show that our objective outperforms the data dependent objectives to fool the learned models. Further, via exploiting simple priors related to the data distribution, our objective remarkably boosts the fooling ability of the crafted perturbations. Significant fooling rates achieved by our objective emphasize that the current deep learning models are now at an increased risk, since our objective generalizes across multiple tasks without the requirement of training data for crafting the perturbations. To encourage reproducible research, we have released the codes for our proposed algorithm\footnotemark{$^\dagger$}.
\end{abstract}

% Note that keywords are not normally used for peer review papers.
}

% make the title area
\maketitle

% To allow for easy dual compilation without having to reenter the
% abstract/keywords data, the \IEEEtitleabstractindextext text will
% not be used in maketitle, but will appear (i.e., to be "transported")
% here as \IEEEdisplaynontitleabstractindextext when the compsoc 
% or transmag modes are not selected <OR> if conference mode is selected 
% - because all conference papers position the abstract like regular
% papers do.
\IEEEdisplaynontitleabstractindextext
% \IEEEdisplaynontitleabstractindextext has no effect when using
% compsoc or transmag under a non-conference mode.

% For peer review papers, you can put extra information on the cover
% page as needed:
% \ifCLASSOPTIONpeerreview
% \begin{center} \bfseries EDICS Category: 3-BBND \end{center}
% \fi
%
% For peerreview papers, this IEEEtran command inserts a page break and
% creates the second title. It will be ignored for other modes.
\IEEEpeerreviewmaketitle
\IEEEraisesectionheading{\section{Introduction}
\label{sec:introduction}}
\makeatletter{\renewcommand*{\@makefnmark}{}
\footnotetext{* denotes equal contribution}\makeatother}
\makeatletter{\renewcommand*{\@makefnmark}{}
\footnotetext{$^\dagger$ https://github.com/val-iisc/gd-uap}\makeatother}

\IEEEPARstart{S}{mall} but structured perturbations to the input, called adversarial perturbations, are shown~(\cite{prsystemsunderattack-pari-2014,evasion-mlkd-2013,adversarialml-acmmm-2011}) to significantly affect the output of machine learning systems. Neural network based models, despite their excellent performance, are  observed~(\cite{intriguing-arxiv-2013,explainingharnessing-arxiv-2014,atscale-arxiv-2016}) to be vulnerable to adversarial attacks. Particularly, Deep Convolutional Neural Networks (CNN) based vision models~(\cite{deepfool-cvpr-2016,universal-cvpr-2017,mopuri-bmvc-2017,advobjdet-arxiv-2017,universalseg-iccv-2017}) can be fooled by carefully crafted quasi-imperceptible perturbations. Multiple hypotheses attempt to explain the existence of adversarial samples, viz. linearity of the models~\cite{explainingharnessing-arxiv-2014}, finite training data~\cite{deeparch-FTML-2009}, etc. More importantly, the adversarial perturbations generalize across multiple models. That is, the perturbations crafted for one model fools another model even if the second model has a different architecture or is trained on a different dataset (\cite{intriguing-arxiv-2013,explainingharnessing-arxiv-2014}). This property of adversarial perturbations enables potential intruders to launch attacks without the knowledge about the target model under attack: an attack typically known as \textit{black-box attack}~\cite{practicalbb-arxiv-2016}. In contrast, an attack where everything about the target model is known to the attacker is called a \textit{white-box attack}. Until recently, all the existing works assumed a threat model in which the adversaries can directly feed input to the machine learning system. However, Kurakin \textit{et al.}~\cite{physicalworld-arxiv-2016} lately showed that the adversarial samples can remain misclassified even if they were constructed in physical world and observed through a sensor (e.g., camera). Given that the models are vulnerable even outside the laboratory setup~\cite{physicalworld-arxiv-2016}, the models' susceptibility poses serious threat to their deploy-ability in the real world (e.g., safety concerns for autonomous driving). Particularly, in case of critical applications that involve safety and security, reliable models need to be deployed to stand against the strong adversarial attacks. Thus, the effect of these structured perturbations has to be studied thoroughly in order to develop dependable machine learning systems. %In this work we consider the objective

Recent work by Moosavi-Dezfooli \textit{et al.}~\cite{universal-cvpr-2017} presented the existence of image-agnostic perturbations, called universal adversarial perturbations (UAP) that can fool the state-of-the-art recognition models on most natural images. %They also demonstrate surprising transferability (ability to fool) for the crafted UAPs across multiple architectures trained on the same target data.
% Their method for crafting the UAPs is based on the DeepFool~\cite{deepfool-cvpr-2016} attacking method. It involves solving a complex optimization problem (eqn.~\ref{eqn:existing}) to design a perturbation. The UAP~\cite{universal-cvpr-2017} procedure utilizes a set of training images to iteratively update the universal perturbation with an objective of changing the predicted label upon addition, for most of the dataset images. Similar to~\cite{universal-cvpr-2017}, Metzen~\textit{et al.}~\cite{universalseg-iccv-2017} proposed UAP for semantic segmentation task. They extended the iterative FGSM~\cite{explainingharnessing-arxiv-2014} attack by Kurakin \textit{et al.}~\cite{physicalworld-arxiv-2016} to change the label predicted at individual pixels and craft the perturbation. They craft the image-agnostic perturbations to fool the system in order to predict a pre-determined target segmentation output.
Their method for crafting the UAPs, based on the DeepFool~\cite{deepfool-cvpr-2016} attacking method, involves solving a complex optimization problem (eqn.~\ref{eqn:existing}) to design a perturbation. The UAP~\cite{universal-cvpr-2017} procedure utilizes a set of training images to iteratively update the universal perturbation with an objective of changing the predicted label upon addition. Similar to~\cite{universal-cvpr-2017}, Metzen~\textit{et al.}~\cite{universalseg-iccv-2017} proposed UAP for semantic segmentation task. They extended the iterative FGSM~\cite{explainingharnessing-arxiv-2014} attack by Kurakin \textit{et al.}~\cite{physicalworld-arxiv-2016} to change the label predicted at each pixel. Additionally, they craft image-agnostic perturbations to fool the system in order to predict a pre-determined target segmentation output.

However, these approaches to craft UAPs~(\cite{universal-cvpr-2017,universalseg-iccv-2017,segmentation-objectdetection-iccv-2017}) have the following important drawbacks:
\begin{itemize}
    \item \textit{Data dependency}: It is observed that the objective presented by~\cite{universal-cvpr-2017} to craft UAP requires a minimum number of training samples for it to converge and craft an image-agnostic perturbation. Moreover, the fooling performance of the resulting perturbation is proportional to the available training data (Figure~\ref{fig:dd-vs-df}). Similarly, the objective for semantic segmentation models (e.g., \cite{universalseg-iccv-2017}) also requires data. Therefore, existing procedures can not craft perturbations when enough data is not provided.
    \item \textit{Weaker black-box performance}: Since information about the target models is generally not available for attackers, it is practical to study the  black-box attacks. Also, black-box attacks reveal the true susceptibility of the models, while white-box attacks provide an upper bound on the achievable fooling. However, the black-box attack of UAP~\cite{universal-cvpr-2017} is significantly weaker than their white-box attack (Table~\ref{tab:dd-vs-df}). Note that, in~\cite{universal-cvpr-2017}, authors have not analyzed the performance of their perturbations in the black-box attack scenario. They have assumed that the training data of the target models is known and have not considered the case in which adversary has access to only a different set of data. This amounts to performing only \textit{semi white-box} attacks. Black-box attacks generally imply (\cite{practicalbb-arxiv-2016}) that the adversary does not have access to (i) the target network architecture (including the parameters), and (ii) a large training dataset. Even in the case of semantic segmentation, since~\cite{universalseg-iccv-2017} work with targeted attacks, they observed that the perturbations do not generalize to other models very well.
    \item \textit{Task specificity}: The current objectives to craft UAPs are task specific. The objectives are typically designed to suit the underlying task at hand since the concept of fooling varies across the tasks. Particularly, for regression tasks such as depth estimation and crowd counting, extending the existing approaches to craft UAPs is non-trivial.
\end{itemize}

%this approach is not suitable for crafting perturbations in most practical scenarios. practical attacks since the training data of the target systems is generally not available for the attacker. Thus, (their) data dependent objective can only reveal the existence of such a direction in the data space that can be exploited to attack the models. Further, the fooling rates achieved by the crafted UAPs may not be representative, rather, they might be treated as an upper bound on the achievable fooling. 

In order to address the above shortcomings and to better analyze the stability of the models, we present a novel data-free objective to craft universal adversarial perturbations, called \textit{GD-UAP}. Our objective is to craft image-agnostic perturbations that can fool the target model without any knowledge about the data distribution, such as, the number of categories, type of data (e.g., faces, objects, scenes, etc.) or the data samples themselves.  Since we do not want to utilize any data samples, instead of an objective that reduces the confidence to the predicted label or flip the predicted label (as in~\cite{deepfool-cvpr-2016, intriguing-arxiv-2013,universal-cvpr-2017,universalseg-iccv-2017}), we propose an objective to learn perturbations that can adulterate the features extracted by the models. Our proposed objective attempts to over-fire the neurons at multiple layers in order to deteriorate the extracted features. During the inference time, the added perturbation misfires the neuron activations in order to contaminate the representations and eventually lead to wrong prediction.

This work extends our earlier conference paper~\cite{mopuri-bmvc-2017}. We make the following new contributions in this paper:
\begin{enumerate}
    \item We propose a novel data-free objective for crafting image-agnostic perturbations.
    \item We demonstrate that our objective is generalizable across multiple vision tasks by extensive evaluation of the crafted perturbations across three different vision tasks covering both classification and regression.
    \item Further, we show that apart from being data-free objective, the proposed method can exploit minimal prior information about the training data distribution of the target models in order to craft stronger perturbations.
    \item We present comprehensive analysis of the proposed objective which includes: (a) a thorough comparison of our approach with the data-dependant counterparts, and (b) evaluation of the strength of UAPs in the presence of various defense mechanisms.
%    \item 
    %the crafted perturbations across three different vision tasks covering both classification and regression tasks.
\end{enumerate}
The rest of this paper is organized as follows: section \ref{sec:relworks} presents detailed account of related works, section~\ref{sec:proposed} discusses the proposed data-free objective to craft image-agnostic adversarial perturbations, section~\ref{sec:expts} demonstrates the effectiveness of \textit{GD-UAP} to craft UAPs across various tasks, section~\ref{sec:anal} hosts a thorough experimental analysis of \textit{GD-UAP} and finally section~\ref{sec:conclu} concludes the paper.

%%%%%%%%%%%%%%%%%%%%%%%%%%%%%%%%%%%%%%%%%%%%%%%%%%%%%%%%%%%%%%%%%%%
%  sec-2 Related Works
%%%%%%%%%%%%%%%%%%%%%%%%%%%%%%%%%%%%%%%%%%%%%%%%%%%%%%%%%%%%%%%%%%%
\section{Related Works}
\label{sec:relworks}
Szegedy \textit{et al.}~\cite{intriguing-arxiv-2013} demonstrated that despite their superior recognition performance, neural networks are susceptible to adversarial perturbations. Subsequently, multiple other works~\cite{robustnessanalysis-arxiv-2015, adversarial2noise-nips-2016,explainingharnessing-arxiv-2014, atscale-arxiv-2016,deepfool-cvpr-2016,easilyfooled-cvpr-2015,hardpositives-cvpr-2016} studied this interesting and surprising property of the machine learning models. Though it is first observed with recognition models, the adversarial behaviour is noticed with models trained on other tasks such as semantic segmentation~\cite{universalseg-iccv-2017,segmentation-detection-iccv-2017}, object detection~\cite{segmentation-detection-iccv-2017}, pose estimation~\cite{cisse2017houdini} and deep reinforcement learning tasks~\cite{adverserl-arxiv-2017}. There exist multiple methods to craft these malicious perturbations for a given data sample. For recognition tasks, they range from performing simple gradient ascent \cite{explainingharnessing-arxiv-2014} on cost function to solving complex optimizations ( \cite{intriguing-arxiv-2013,deepfool-cvpr-2016,measuringrobustness-nips-2016} ). Simple and fast methods such as FGSM~\cite{explainingharnessing-arxiv-2014} find the gradient of loss function to determine the adversarial perturbation. An iterative version of this attack presented in~\cite{physicalworld-arxiv-2016} achieves better fooling via performing the gradient ascent multiple times. On the other hand, complex approaches such as~\cite{deepfool-cvpr-2016} and \cite{intriguing-arxiv-2013} find minimal perturbation that can move the input across the learned classification boundary in order to flip the predicted label. More robust adversarial attacks have been proposed recently that transfer to real world~\cite{physicalworld-arxiv-2016} and are invariant to general image transformations~\cite{athalye-arxiv-2017}. 

Moreover, it is observed that the perturbations exhibit transferability, that is, perturbations crafted for one model can fool other models with different architectures and different training sets as well~(\cite{intriguing-arxiv-2013,explainingharnessing-arxiv-2014}). Further, Papernot \textit{et al.}~\cite{practicalbb-arxiv-2016} introduced a practical attacking setup via model distillation to understand the \textit{black-box} attack. Black-box attack assumes no information about the target model and its training data. They proposed to use a target model's substitute to craft the perturbations.

The common underlying aspect of all these techniques is that they are intrinsically data dependent. The perturbation is crafted for a given data sample independently of others. However, recent works by Moosavi-Dezfooli \textit{et al.}~\cite{universal-cvpr-2017} and Metzen \textit{et al.}~\cite{universalseg-iccv-2017} showed the existence of input-agnostic perturbations that can fool the models over multiple images. In~\cite{universal-cvpr-2017}, authors proposed an iterative procedure based on Deepfool attacking~\cite{deepfool-cvpr-2016} method to craft a universal perturbation to fool classification models. Similarly, in~\cite{universalseg-iccv-2017}, authors craft universal perturbations that can affect target segmentation output. However, both these works optimize for different task specific objectives. Also, they require training data to craft the image-agnostic perturbations. Unlike the existing works, our proposed method \textit{GD-UAP} presents a data-free objective that can craft perturbations without the need for any data samples. Additionally, we introduce a generic notion of fooling across multiple computer vision tasks via over-firing the neuron activations. Particularly, our objective is generalizable across various vision models in spite of differences in terms of architectures, regularizers, underlying tasks, etc.

\section{Proposed Approach}
\label{sec:proposed}
%%%%%%%%%%%%%%%%%%%%%%%%%%%%%%%%%%%%%%%%%%%%%%%%%%%%%%%%%%%%%%%%%%%
\begin{figure}
\centering
\includegraphics[width=0.45\textwidth]{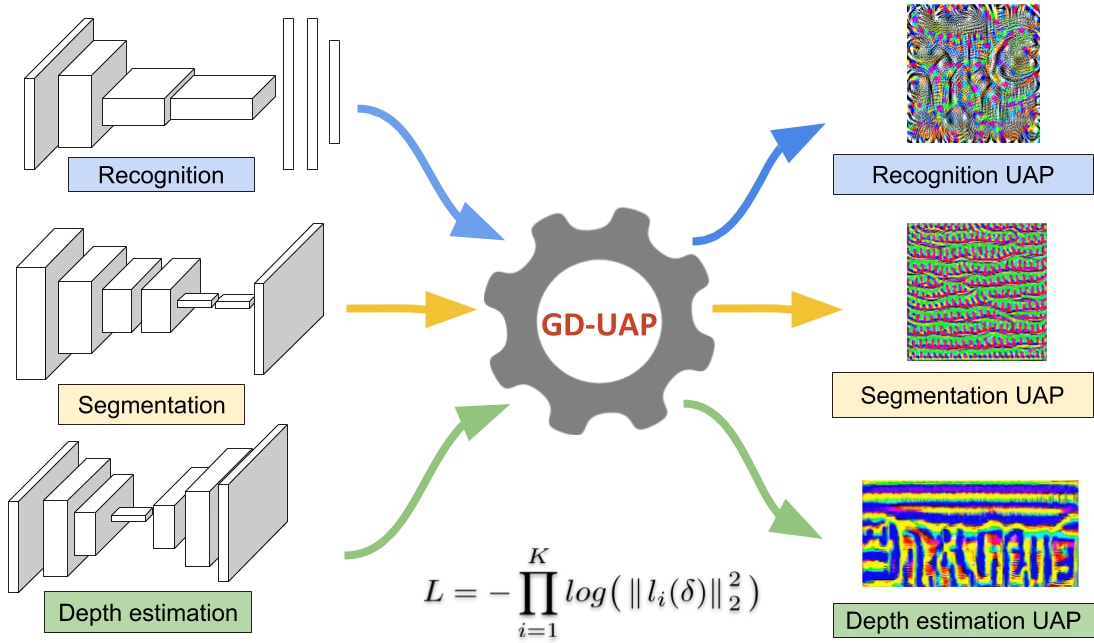} %GD-UAP-overview-final.pdf
\caption{Overview of the proposed generalized objective to craft ``Image agnostic" (Universal) Adversarial Perturbations for a given target CNN. Input to our method is a task specific target CNN. The proposed objective, which is independent of the underlying task, then crafts UAPs without utilizing any data samples. The crafted UAPs are transferable to other models trained to perform the same underlying task as the target CNN.}
\label{fig:gduap-overview}
\end{figure}
%  sec-3 Proposed Approach
%%%%%%%%%%%%%%%%%%%%%%%%%%%%%%%%%%%%%%%%%%%%%%%%%%%%%%%%%%%%%%%%%%%
%In this section, we discuss the proposed data-free objective to craft UAPs in detail.
First, we introduce the notation followed throughout the paper. $\mathcal{X}$ denotes the distribution of images in $\mathbb{R}^d$. $f$ denotes the function learned by the CNN that maps an input image $x \sim \mathcal X$ to its output $f(x)$. %ote that the output is task dependent, for example, it is a label for object recognition and segmentation map for semantic segmentation.
 $\delta$ denotes the image-agnostic perturbation learned by our objective. Similar to input $x$, $\delta$ also belongs to $\mathbb{R}^d$. Though the proposed objective is task independent, for ease of understanding we explain the proposed approach in the context of object recognition. Note that the proposed \textit{objective} is generalizable across multiple vision tasks to craft \textit{task specific}, image-agnostic adversarial perturbations. Insterestingly, these crafted task specific perturbations exhibit cross model generalizability. Figure~\ref{fig:gduap-overview} presents the proposed generalizable approach to learn task specific, image agnostic perturbations in data-free scenario.

\subsection{Data-free objective for fooling}
\label{subsec:data-free-fooling}
The objective of our paper is to craft an image-agnostic perturbation $\delta \in \mathbb{R}^d$ that fools the CNN $f$ for images from the target distribution $\mathcal{X}$ without utilizing any samples from it. %In other words, we seek a universal adversarial perturbation $\delta$ that significantly alters the prediction of the CNN $(f)$.
That is, we synthesize a $\delta$ such that

\begin{equation}
    f(x+\delta) \neq f(x), \:\:\:\text{for $x \sim \mathcal X$}.
\end{equation}

The pixel intensities of $\delta$ are restricted by an imperceptibility constraint. Typically, it is realized as a max-norm constraint in terms of $l_{\infinity}$ or $l_2$ norms (e.g.~\cite{universal-cvpr-2017,mopuri-bmvc-2017,universalseg-iccv-2017,explainingharnessing-arxiv-2014}). In this paper, for all our analysis we impose $l_{\infinity}$ norm. Thus, the aim is to find a $\delta$ such that
\begin{align}
\label{eqn:existing}
\begin{split}
  f(x+\delta) &\neq f(x), \:\:\:\text{for $x \in \mathcal X$}; \\
  \Vert\: \delta\: \Vert_{\infinity} &< \xi.
\end{split}
\end{align}

However, the focus of the proposed work is to craft $\delta$  without requiring any data samples. The data-free nature of our approach prohibits us from utilizing  eqn.~\ref{eqn:existing} for learning $\delta$, as we do not have access to data $x$.
% That is, our approach does not contain the data term $x$ in the proposed objective. 
Therefore, we instead propose to fool the CNN by contaminating the extracted representations of the input at multiple layers of the architecture. In other words, as opposed to the typical ``flipping the label" objective, we attempt to ``over-fire" the features extracted at multiple layers. That is, we craft a perturbation $\delta$ such that it leads to additional activation firing at each layer and thereby misleading the features (filters) at the following layer. The accumulated effect of the contamination eventually leads the CNN to misclassify.

The perturbation essentially causes filters at a particular layer to spuriously fire and extract inefficacious information. Note that in the presence of data (during attack), in order to mislead the activations from retaining useful discriminative information, the perturbation $(\delta)$ has to be highly effective. Also, the imperceptibility constraint (second part of eqn.~\ref{eqn:existing}) on $\delta$ makes it more challenging.

Hence without utilizing any data $x$, we seek an image-agnostic perturbation $\delta$ that can produce maximal spurious activations at each layer of a given CNN. In order to craft such a $\delta$ we start with a random perturbation and optimize for the following objective:
\begin{align}
\label{eqn:data-free-objective}
\begin{split}
  Loss = - \log \left( \prod\limits_{i=1}^K  \Vert l_i(\delta)\Vert_2 \right), \:\:\:\:\:\: \text{such that} \:\:\:\:\:\:
  \Vert\: \delta\: \Vert_{\infinity} < \xi.
\end{split}
\end{align}

where $l_i(\delta)$ is the activation in the output tensor (after the non-linearity) at layer $i$ when $\delta$ is fed to the network $f$. $K$ is the number of layers in $f$ at which we maximize the activations caused by $\delta$, and $\xi$ is the max-norm limit on $\delta$.

The proposed objective computes product of activation magnitude at all the individual layers. We observed product resulting in stronger $\delta$ than other forms of aggregation (e.g.\ sum). %This is understandable, since product can force the individual activations to rise in order for the loss to reduce.
 To avoid working with extreme values ($\approx 0$), we apply log on the product. Note that the objective is open-ended as there is no optimum value to reach. We would ideally want $\delta$ to cause as much strong disturbance at all the layers as possible, within the imperceptibility constraint. More discussion on the motivation and working of the proposed objective is presented in Section~\ref{sec:anal}.

\subsection{Implementation Details}
\label{subsec:implementation-details}
We begin with a target network $f$ which is a trained CNN whose parameters are frozen and a random perturbation $\delta$. We then perform the proposed optimization to update $\delta$ for causing strong activations at multiple layers in the given network. Typically, it is considered that the convolution $(conv)$ layers learn information-extracting features which are then classifed by a series of $fc$ layers. Hence, we optimize our objective only at all the $conv$ layers. This was empirically found to be more effective than optimizing at all layers as well. %Therefore, we restrict the optimization to feature extraction layers.
%We typically consider all the convolution $(conv)$ layers before the fully connected $(fc)$ layers. This is because, the $conv$ layers are generally considered to learn required features to extract information over which a series of $fc$ layers perform classification. Also, we empirically found that it is effective to optimize at $conv$ layers. %Therefore, we restrict the optimization to feature extraction layers.
 In case of advanced architectures such as GoogLeNet~\cite{googlenet-arxiv-21014} and ResNet~\cite{resnet-2015}, we optimize at the last layers of all the inception (or residual) blocks and the independent $conv$ layers. We observed that optimizing at these layers results in $\delta$ with a fooling capacity similar to the one resulting from optimizing at all the intermediate layers as well (including the $conv$ layers within the inception/residual blocks). However, since optimizing at only the last layers of these blocks is more efficient, we perform the same.

Note that the optimization updates only the perturbation $\delta$, not the network parameters. Additionally, no image data is involved in the optimization. We update $\delta$ with the
gradients computed for loss in eqn.~(\ref{eqn:data-free-objective}) iteratively till the fooling performance of the learned $\delta$ gets saturated on a set of validation images. In order to validate the fooling performance of the learned $\delta$, we compose an unrelated substitute dataset $(D)$. Since our objective is not to utilize data samples from the training dataset, we randomly select $1,000$ images from a substitute dataset to serve as validation images. It is a reasonable assumption for an attacker to have access to $1,000$ unrelated images. For crafting perturbations to object recognition models trained on ILSVRC dataset~\cite{imagenet-ijcv-2015}, we choose random samples from Pascal VOC-2012~\cite{pascal-voc-2012} dataset. Similarly, for semantic segmentation models trained on Pascal VOC~\cite{pascal-voc-2011,pascal-voc-2012}, we choose validation samples from ILSVRC~\cite{imagenet-ijcv-2015}, for depth estimation models trained on KITTI dataset~\cite{Geiger2012CVPR} we choose samples from Places-205~\cite{places2-arxiv-2016} dataset.

\subsection{Exploiting additional priors}
\label{subsec:additional-priors}
Though \textit{GD-UAP} is a data-free optimization for crafting image-agnostic perturbations, it can exploit simple additional priors about the data distribution $\mathcal{X}$. In this section we demonstrate how \textit{GD-UAP} can utilize simple priors such as (i) mean value and dynamic range of the input, and (ii) target data samples.

\subsubsection{Mean and dynamic range of the input}
\label{subsubsec:range-prior}
Note that the proposed optimization (eqn.~(\ref{eqn:data-free-objective})) does not consider any information about $\mathcal{X}$. %not even the dynamic range of the input data.
We present only the norm limited $\delta$ as input and maximize the resulting activations. Hence, during the optimization, input to the target CNN has a dynamic range of $[-\xi,\:\xi]$ $(\xi=10)$. However, during the inference time, input lies in $[0,\: 255]$ range. Therefore, it becomes very challenging to learn perturbations that can affect the neuron activations in the presence of strong (an order higher) input signal $x$. Hence, in order to make the learning easier, we may provide this useful information about the data $(x \in \mathcal{X})$, and let the optimization better explore the space of perturbations. Thus, we slightly modify our objective to craft $\delta$ relative to the dynamic range of the data. We create pseudo data $d$ via randomly sampling from a Gaussian distribution whose mean $(\mu)$ is equal to the mean of training data and variance $(\sigma)$ is such that $99.9\%$ of the samples lie in $[0,\: 255]$, the dynamic range of input. Thus, we solve for the following loss:

\begin{align}
\label{eqn:gaussian-nosie-objective}
\begin{split}
  Loss & = -\sum_{d \sim \mathcal{N}(\mu,\sigma) }{ \log \left( \prod\limits_{i=1}^K  \Vert l_i(d+\delta)\Vert_2 \right)}, \\  \text{such that\:\:\:} &
  \Vert \delta\: \Vert_{\infinity} < \xi.  %\text{ and } d \sim \mathcal{N}(\mu,\sigma).
\end{split}
\end{align}
Essentially, we operate the proposed optimization in a subspace closer to the target data distribution $\mathcal{X}$. In other words, $d$ in eqn.~(\ref{eqn:gaussian-nosie-objective}) acts as a place holder for the actual data and helps to learn perturbations which can over-fire the neuron activations in the presence of the actual data. A single Gaussian sample with twice the size of the input image is generated. Then, random crops from the Gaussian sample, augmented with simple techniques such as random cropping, blurring, and rotation are used for the optimization.

\subsubsection{Target data samples}
\label{subsubsec:data-prior}
Now, we modify our data-free objective to utilize samples from the target distribution $ \mathcal{X} $ and improve the fooling ability of the crafted perturbations. Note that in the case of data availability, we can design direct objectives such as reducing confidence for the predicted label or changing the predicted label, etc. However, we investigate if our data-free objective of over-firing the activations, though is not designed to utilize data, crafts better perturbations when data is presented to the optimization. Additionally, our objective does not utilize data to manipulate the predicted confidences or labels. Rather, the optimization benefits from prior information about the data distribution such as the dynamic range, local patterns, etc., which can be provided through the actual data samples. Therefore, with minimal data samples we solve for the following optimization problem

\begin{align}
\label{eqn:data-prior}
\begin{split}
  Loss & = - \sum_{ x \sim \mathcal{X}}{\log \left( \prod\limits_{i=1}^K  \Vert l_i(x+\delta)\Vert_2 \right)}, \\  \text{such that\:\:\:} &
  \Vert \delta\: \Vert_{\infinity} < \xi.
\end{split}
\end{align}
Presenting data samples to the optimization procedure is
 a natural extension to presenting the dynamic range of the target data alone (section~\ref{subsubsec:range-prior}). In this case, we utilize a subset of training images on which the target CNN models are trained (similar to~\cite{universal-cvpr-2017,universalseg-iccv-2017}). %----------------------s--------------------
\subsection{Improved Optimization}
\label{subsec:improved-optimization}
In this subsection, we present improvements to the optimization process presented in our earlier work~\cite{mopuri-bmvc-2017}. We observe that the proposed objective quickly accumulates $\delta$ beyond the imposed max-norm constraint $(\xi)$. Because of the clipping performed after each iteration, the updates after $\delta$ reaches the constraint are futile . To tackle this saturation, $\delta$ is re-scaled to half of its dynamic range (i.e.\ $[-5,\: 5]$). % in regular time intervals of $300$ iterations. 
Not only does the re-scale operation allow an improved utilization of the gradients, it also retains the pattern learnt in the optimization process till that iteration. 

In our previous work~\cite{mopuri-bmvc-2017}, the re-scale operation is done in a regular time interval of $300$ iterations. Though this re-scaling helps to learn better $\delta$, it is inefficient since it performs blind re-scaling without verifying the scope for updating $\delta$. This is specially harmful later in the learning process, when the perturbation may not be re-saturated in $300$ iterations.% as the learning progresses, magnitude of updates decreases and during the interval of $300$ iterations, the values of $\delta$ might not reach the extreme values of $\pm 10$. Projecting the $\delta$ by re-scaling can badly affect the learning.

Therefore, we propose an adaptive re-scaling of $\delta$ based on the rate of saturation (reaching the extreme values of $\pm 10$) in its pixel values. During the optimization, at each iteration we compute the proportion $(p)$ of the pixels in $\delta$ that reached the max-norm limit $\xi$. As the learning progresses, more number of pixels reach the max-norm limit and because of the clipping, eventually get saturated at $\xi$. Hence, the rate of increase in $p$ decreases as $\delta$ saturates. We compute the rate of saturation, denoted as $S$, of the pixels in $\delta$ after each iteration during the training. For consecutive iterations, if increase in $p$ is not significant (less than a pre-defined threshold $\theta$), we perform a re-scaling to half the dynamic range. %Note that the proposed criterion for re-scaling is similar to the typical usage of validation performance to stop training. %
We observe that this adaptive re-scaling consistently leads to better learning.

\subsection{Algorithmic summarization}
\label{subsec:algo}
In this subsection, for the sake of brevity we summarize the proposed approach in the form of an algorithm. Algorithm~\ref{algo:guap} presents the proposed optimization as a series of steps. Note that it is a generic form comprising of all the three variations including both data-free and with prior versions.

For ease of reference, we repeat some of the notation. $F_t$ is the fooling rate at iteration $t$, $l_i(x)$ is the activation caused at layer $i$ of the CNN $f$ for an input $x$, $\eta$ is the learning rate used for training, $\Delta$ is the gradient of the loss with respect to the input $\delta$, $S_t$ is the rate of saturation of pixels in the perturbation $\delta$ at iteration $t$, $\theta$ is the threshold on the rate of saturation, $F_t$ is the fooling rate, $H$ is the patience interval of validation for verifying the convergence of the proposed optimization.

\begin{algorithm}
  \KwData{Target CNN $f$, data $g$. Note that $g=0$ for data-free case, $g= d \sim \mathcal{N}(\mu,\sigma)$ for range prior case, and $g =x$  for training data samples case.}
  \KwResult{Image-agnostic adversarial perturbation $\delta$.}
  Randomly initialize $\delta_0 \sim \mathcal{U}[-\xi,\: \xi]$ \\
  $t=0$\\
  $F_t=0$ \\
  \Do{$ F_t < \text{min. of } \{F_{t-H},F_{t-H+1}\ldots F_{t-1}\}$ }{
    $t \leftarrow t+1$ \\
    Compute $l_{i}(g+\delta)$ \\%via feed forward $(g+\delta)$ through the target CNN $f$ \\
    Compute loss = $- \sum{\log \left( \prod\limits_{i=1}^K  \Vert l_i(g+\delta)\Vert_2 \right)}$ \\%according to one of the equations \ref{eqn:data-free-objective} or \ref{eqn:gaussian-nosie-objective} or \ref{eqn:data-prior} \\
    Update\footnote{} $\delta_t : \delta_{t} \leftarrow \delta_{t-1} - \eta \Delta$\\
    Compute the rate of saturation $S_t$ in the $\delta_t$ pixels\\
    \If{$ S_t < \theta  $}{
      $\delta_t \leftarrow \delta_t/2\;$ %\% Adaptive re-scaling
      }
    Compute $F_t$ of $\delta_t$ on substitute dataset $D$ \\
  }
  $j \leftarrow \text{argmax. of } \{F_{t-H},F_{t-H+1}\ldots F_{t-1}\}$ \\
  Return $\delta_{j}$
  \caption{Algorithm summarizing our approach to craft image-agnostic adversarial perturbations via data-free objective and exploiting various data priors.}
  \algorithmfootnote{$^{1}$Note that the generic update equation 8 is only representative and not the exact equation implemented.}
\label{algo:guap}
\end{algorithm}

\subsection{Generalized Fooling Rate (GFR)}
\label{subsec:GFR}
While the notion of `fooling' has been well defined for the task of image recognition, for other tasks it is unclear. Hence, in order to provide an interpretable metric to measure `fooling', we introduce Generalized Fooling Rate (\textit{GFR}), making it independent of the task, and dependent on the metric being used for evaluating the model's performance.

%Consider the task of image recognition. For this task, fooling rate for any perturbation is defined as the $\%$ of data samples, for which the labels are changed due to the perturbation. i.e., the $\%$ of data samples for which the output label before and after the adversarial attack are different. If the fooling rate is $x\%$, then for $(100-x)\%$ of the data samples, the label before and after the adversarial attack remains the same. %If we consider the labels of clean samples as ground truth, and labels of perturbed samples as predicted labels, $(100-x)\%$ is same as the Top-1 accuracy of the model. This leads us to introduce the following definition for the Generalized Fooling Rate.

Let $M$ be a metric for measuring the performance of a model for any task, where the range of $M$ is $[0,R]$. Let the metric take two inputs $\hat{y}$ and $y$, where $\hat{y}$ is the predicted output and $y$ is the ground truth output, such that the performance of the model is measured as $M(\hat{y},y)$. Let $\hat{y_{\delta}}$ be the output of the model when the input is perturbed with a perturbation $ \delta $. Then, the Generalized Fooling Rate with respect to measure $M$ is defined as:
\begin{align}
\label{eqn:GFR}
\begin{split}
  GFR(M) & = \frac{R - M(\hat{y_{\delta}},\hat{y})}{R}.
\end{split}
\end{align}

This definition of Generalized Fooling rate (\textit{GFR}) has the following benefits:
\begin{itemize}
\item \textit{GFR} is a natural extension of `fooling rate' defined for image recognition, where the fooling rate can be written as $GFR(Top1) = 1 - Top1(\hat{y_{\delta}},\hat{y})$, where $Top1$ is the Top-1 Accuracy metric. %(ref eq.~\ref{eqn:GFR}).
\item Fooling rate should be a measure of the change in model's output caused by the perturbation. Being independent of the ground truth $y$, and dependant only on $\hat{y_{\delta}}$ and $\hat{y}$, \textit{GFR} primarily measures the change in the output. A poorly performing model which however is very robust to adversarial attacks will show very poor \textit{GFR} values, highlighting its robustness.% For example, GFR will be equal to $0\%$ if there is no change in the output, and in extreme cases it can reach $100\%$, for a given task. Hence it can be compared across tasks. 
\item \textit{GFR} measures the performance of a perturbation in terms of the damage caused to a model with respect to a metric. This is an important feature, as tasks such as depth estimation have multiple performance measures, where some perturbation might cause harm only to some of the metrics while leaving other metrics unaffected.
\end{itemize}

For all the tasks considered in this work, we report \textit{GFR} with respect to a metric as a measure of `fooling'.
%\begin{algorithm}
%\caption{Algorithm summarizing our approach to craft image-agnostic adversarial perturbations via data-free objective and exploiting various data priors.}\label{algo:guap}
%\hspace*{\algorithmicindent} \textbf{Input}{: Target CNN $f$, data $g$. Note that $g=0$ for data-free case, $g= d \sim \mathcal{N}(\mu,\sigma)$ for range prior case, and $g =x$  for target data samples case} \\
%\hspace*{\algorithmicindent} \textbf{Output}{: Image-agnostic adversarial perturbation $\delta$.} 
%\begin{algorithmic}[1]
%\Procedure{G-UAP}{}
%\State Randomly initialize $\delta \sim \mathcal{U}[-\xi,\: \xi]$  
%\Statex \emph{Begin loop}:
%\bindent
%\State Compute $l_{i}(g+\delta)$ via feed forward $(g+\delta)$ through the target CNN $f$
%\State Compute the suitable loss according to one of the equations %\ref{eqn:data-free-objective}, \ref{eqn:gaussian-nosie-objective} or %\ref{eqn:data-prior}
%\State Update $\delta$ with the computed gradients
%\State Compute the rate of saturation $S$ in the $\delta$ pixels
%\Statex \emph{Adaptive re-scaling of $\delta$}:
%\If {$ Th > S$ } 
%$\delta \leftarrow \delta/2$%\Return false
%\EndIf
%\Statex \emph{Stopping criterion}:
%\State Compute the fooling performance $F_t$ of $\delta$ on a held-out set
%\If {$ F_t < \text{min. of} \{F_{t-H},F_{t-H+1}\ldots F_{t-1}\}$ } 
%return $\delta$
%\EndIf
%\Statex \emph{End loop}
%\eindent
%\end{algorithmic}
%\end{algorithm}
%++++++++++++++++++++++++++++++++++++++++++++++++++++++++++
%+++++++++        EXPERIMENTS SECTION        ++++++++++++++
%++++++++++++++++++++++++++++++++++++++++++++++++++++++++++

\section{GD-UAP: Effectiveness across tasks}
\label{sec:expts}

In this section, we present the experimental evaluation to demonstrate the effectiveness of \textit{GD-UAP}. We consider three different vision tasks to demonstrate the generalizability of our objective, namely, object recognition, semantic segmentation and unsupervised monocular depth estimation. Note that the set of applications include both classification and regression tasks. Also, it has both supervised and unsupervised learning setups, and various archtectures such as fully convolutional networks, and encoder-decoder networks. We explain each of the tasks separately in the following subsections.

%%%%%%%%%%% Table-1 Fooling rates (data-free, range prior ILSVRC  starts %%%%%%%%%
\begin{table}[t]
\caption{Fooling rates for \textit{GD-UAP} perturbations learned for object recognition on ILSVRC dataset~\cite{imagenet-ijcv-2015}. Each row of the table shows fooling rates for perturbation learned on a specific target model when attacking various other models (columns). These rates are obtained by \textit{GD-UAP} objective with range prior (sec.~\ref{subsubsec:range-prior}). Diagonal rates indicate white-box attack scenario and off-diagonal ones represent black-box attack scenario.}
\centering
\label{tab:fooling-transfer}
\setlength\tabcolsep{1.5pt} 
\begin{tabular}{@{}lccccccc@{}}
%\small
\toprule
Model        & CaffeNet       & VGG-F          & GoogLeNet      & VGG-16         & VGG-19         & Resnet-152     \\ \midrule
CaffeNet     & \textbf{87.02} & 65.97          & 49.40          & 50.46          & 49.92          & 38.57          \\
VGG-F        & 59.89          & \textbf{91.91} & 52.24          & 51.65          & 50.63          & \textbf{40.72}         \\
GoogLeNet    & 44.70          & 46.09          & \textbf{71.44} & 37.95          & 37.90          & 34.56         \\
VGG-16       & 50.05          & 55.66          & 46.59          & \textbf{63.08} & 56.04          & 36.84         \\
VGG-19       & 49.11          & 53.45          & 40.90          & 55.73          & \textbf{64.67} & 35.81         \\
Resnet-152   & 38.41          & 37.20          & 33.22          & 27.76          & 26.52          & 37.3 \\ \bottomrule
\end{tabular}
\end{table}
%%%%%%%%%%% Table-1 Fooling rates (data-free, range prior ILSVRC  ends %%%%%%%%%

% ######           Fig-1 begins       #####################################
\begin{figure}[]
\centering
\noindent\begin{minipage}{\columnwidth}
  \centering
  \begin{minipage}{.15\textwidth}
  	\centering
    \includegraphics[width=\linewidth]{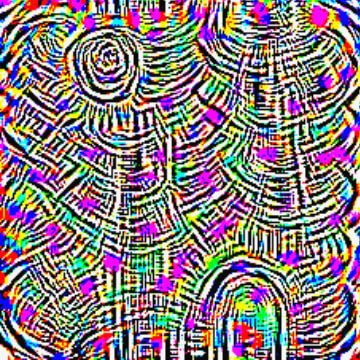}\
   \scriptsize{CaffeNet}
  \end{minipage}
   \begin{minipage}{.15\textwidth}
   	\centering
    \includegraphics[width=\linewidth]{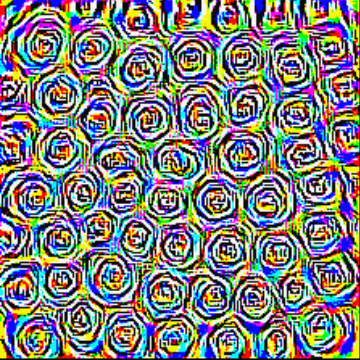}\
    \scriptsize{VGG-F}
  \end{minipage}
  \begin{minipage}{.15\textwidth}
   \centering
    \includegraphics[width=\linewidth]{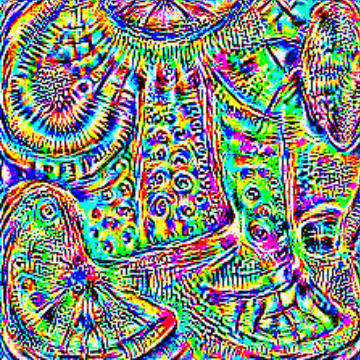}\
   \scriptsize{Googlenet}
  \end{minipage}
  \begin{minipage}{.15\textwidth}
  	\centering
    \includegraphics[width=\linewidth]{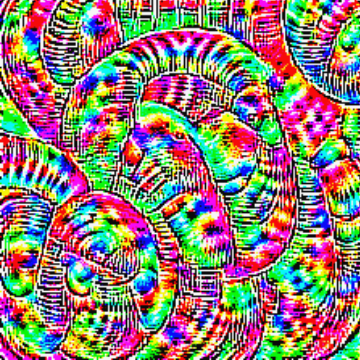}\
    \scriptsize{VGG-$16$}
  \end{minipage}
  \begin{minipage}{.15\textwidth}
  \centering
    \includegraphics[width=\linewidth]{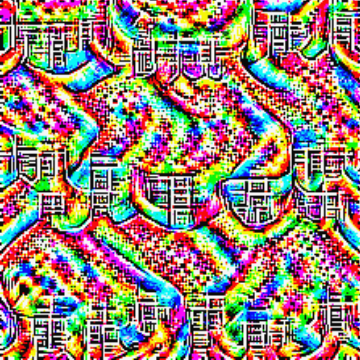}\
    \scriptsize{VGG-$19$}
  \end{minipage}
   \begin{minipage}{.15\textwidth}
   \centering
    \includegraphics[width=\linewidth]{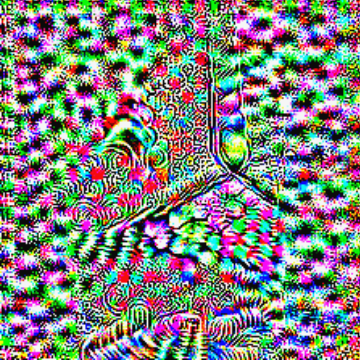}\
    \scriptsize{ResNet-$152$}
  \end{minipage}
    \vspace{0.002\textwidth}
\end{minipage}
%\vspace{.01cm}
\caption{Universal adversarial perturbations crafted by \textit{GD-UAP} objective for multiple models trained on ILSVRC~\cite{imagenet-ijcv-2015} dataset. Perturbations were crafted with $\xi=10$ using the range prior (sec.~\ref{subsubsec:range-prior}).Images are best viewed in color.}
\label{fig:data-free-sample-perturbations-l2}
\end{figure}

% ######           Fig-1 ends       #####################################
% ######           Fig-3 begins       #####################################
\begin{figure}[ht]
\centering
\scriptsize{
\noindent\begin{minipage}{\columnwidth}
  \centering
  \begin{minipage}{.15\textwidth}
  	\centering
    \includegraphics[width=\linewidth]{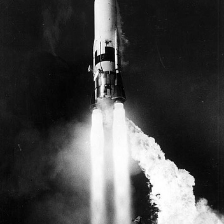}\
    Missile
  \end{minipage}
   \begin{minipage}{.15\textwidth}
   	\centering
    \includegraphics[width=\linewidth]{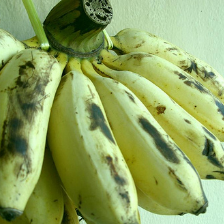}\
    Banana
  \end{minipage}
  \begin{minipage}{.15\textwidth}
  	\centering
    \includegraphics[width=\linewidth]{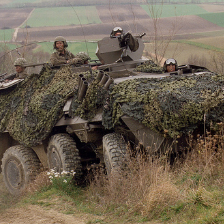}\
    Tank
  \end{minipage}
  \begin{minipage}{.15\textwidth}
  	\centering
    \includegraphics[width=\linewidth]{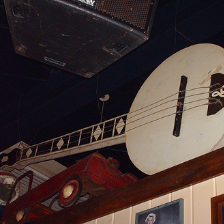}\
    Banjo
  \end{minipage}
   \begin{minipage}{.15\textwidth}
   	\centering
    \includegraphics[width=\linewidth]{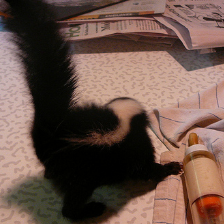}\
    Skunk
  \end{minipage}
  \begin{minipage}{.15\textwidth}
  	\centering
    \includegraphics[width=\linewidth]{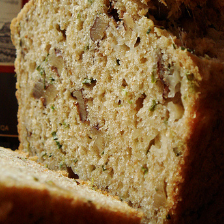}\
     French loaf
  \end{minipage}
  \vspace{0.002\textwidth}
\end{minipage}
\noindent\begin{minipage}{\columnwidth}
  \centering
  \begin{minipage}{.15\textwidth}
  \centering
    \includegraphics[width=\linewidth]{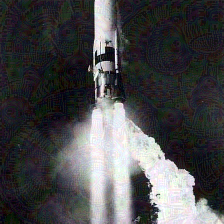}\
    {\color{red} African\\ chamaelon}
  \end{minipage}
   \begin{minipage}{.15\textwidth}
   \centering
    \includegraphics[width=\linewidth]{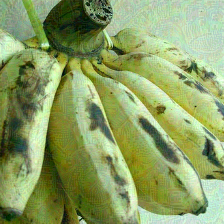}\
    {\color{red} Custard\\ Apple}
  \end{minipage}
  \begin{minipage}{.15\textwidth}
  	\centering
    \includegraphics[width=\linewidth]{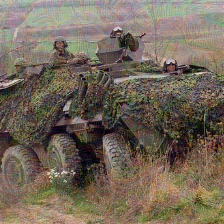}\
    {\color{red} Half\\ track}
  \end{minipage}
  \begin{minipage}{.15\textwidth}
  \centering
    \includegraphics[width=\linewidth]{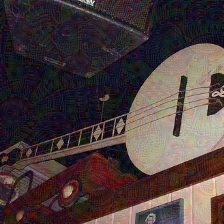}\
     {\color{red}\\Vault }
  \end{minipage}
   \begin{minipage}{.15\textwidth}
   \centering
    \includegraphics[width=\linewidth]{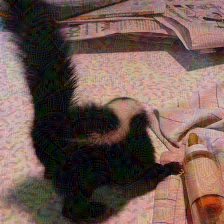}\
    {\color{red} Jigsaw\\ puzzle}
  \end{minipage}
  \begin{minipage}{.15\textwidth}
  	\centering
    \includegraphics[width=\linewidth]{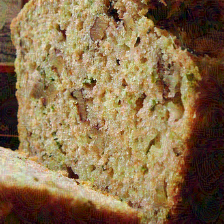}\
    {\color{red} Normal\\ chiton}
  \end{minipage}
    \vspace{0.002\textwidth}
\end{minipage}
}
\vspace{0.1cm}
\caption{Sample original and adversarial image pairs from ILSVRC validation set generated for VGG-19. First row shows original images and corresponding predicted labels, second row shows the corresponding perturbed images along with their predictions.}
\label{fig:data-free-sample-images-l2}
\end{figure}
% ######           Fig-3 ends         #####################################

%In this section, we present the experimental evaluation to demonstrate the effectiveness of the proposed data-free objective. We consider three different vision tasks to demonstrate the generalizability of our objective, namely, object recognition, semantic segmentation and unsupervised monocular depth estimation. Note that the set of applications include both classification and regression tasks. Also, it has both supervised and unsupervised learning setups. We explain each of the tasks separately in the following subsections.

For all the experiments, the ADAM~\cite{KingmaB14_ADAM} optimization algorithm is used with the learning rate of $0.1$. The threshold $\theta$ for the rate of saturation $S$ is set to $10^{-5}$ and $\xi$ value of $10$ is used. Validation fooling rate $F_t$ is measured on the substitute dataset $D$ after every $200$ iterations only when the threshold of the rate of saturation is crossed. If it is not crossed, $F_t$ is measured after every $400$ iterations. Note that the algorithm specific hyper-parameters are \textbf{not} changed across tasks or across priors. 

For all the following experiments, perturbation crafted with different priors are denoted as $P_{NP}, P_{RP},\text{and} P_{DP}$ for the \textit{No prior, Range prior}, and \textit{Data prior} scenario respectively. To emphasize the effectiveness of the proposed objective, we present fooling rates obtained by a random baseline perturbation. Since our learned perturbation is norm limited by $\xi$, we sample random $\delta$ from $\mathcal{U}[-\xi, \xi]$ and compute the fooling rates. In all tables in this section, the `Baseline' column refers to this perturbation.%{\color{blue} Also, we observe that our optimization always converges. This is because the proposed method optimizes for an open-ended objective (activation can be as high as possible). Convergence translates to a state where the activations $(l_i)$ for the perturbation $(\delta)$ saturate to a relatively high value.}

\subsection{Object Recognition}
\label{subsec:object-recognition}
We utilized models trained on ILSVRC~\cite{imagenet-ijcv-2015} and Places-$205$\cite{places2-arxiv-2016} datasets, viz. CaffeNet~\cite{caffe-arxiv-2014}, VGG-F~\cite{vggf-bmvc-2014}, Googlenet~\cite{googlenet-arxiv-21014}, VGG-$16$~\cite{vgg-arxiv-2014}, VGG-$19$~\cite{vgg-arxiv-2014}, ResNet-$152$~\cite{resnet-2015}. For all experiments, pretrained models are used whose weights are kept frozen throughout the optimization process.% Since our approach does not involve training the models, for all the experiments we work with available trained models.
 Also, in contrast to UAP~\cite{universal-cvpr-2017}, we do not use training data in the data-free scenario (sec.~\ref{subsec:data-free-fooling} and sec.~\ref{subsubsec:range-prior}). However, as explained earlier, we use $1,000$ images randomly chosen from Pascal VOC-$2012$~\cite{pascal-voc-2012} training images as validation set ($D$ in Algorithm~\ref{algo:guap}) for our optimization. Also, in case of exploiting additional data prior (sec.~\ref{subsubsec:data-prior}), we use limited data from the corresponding training set. For the final evaluation on ILSVRC of the crafted UAPs, $50,000$ images from the validation set are used. Similarly, for Places-$205$ dataset, $20,500$ images from the validation set are used.

\subsubsection{Fooling performance of the data-free objective}
\label{subsubsec:fooling-data-free}
Table~\ref{tab:fooling-transfer} presents the fooling rates achieved by our objective on various network architectures. Fooling rate is the percentage of test images for which our crafted perturbation $\delta$ successfully changed the predicted label. Using the terminology introduced in sec.~\ref{subsec:GFR}, fooling rate can also be written as $GFR(Top1)$. Higher the fooling rate, greater is the perturbation's ability to fool and lesser is the the classifier's robustness. Fooling rates in Table~\ref{tab:fooling-transfer} are obtained using the mean and dynamic range prior of the training distribution (sec.~\ref{subsubsec:range-prior}). Each row in the table indicates one target model employed in the learning process and the columns indicate various models attacked using the learned perturbations. The diagonal fooling rates indicate the \textit{white-box attacking}, where all the information about the model is known to the attacker. The off-diagonal rates indicate \textit{black-box attacking}, where no information about the model under attack is revealed to the attacker. However, the dataset over which both the models (target CNN and the CNN under attack) are trained is same. Our perturbations cause a mean white-box fooling rate of $69.24\%$ and a mean black-box fooling rate of $45.13\%$. Given the data-free nature of the optimization, these fooling rates are alarmingly significant. The high fooling rates achieved by the proposed approach can adversely affect the real-world deploy-ability of these models.

Figure~\ref{fig:data-free-sample-perturbations-l2} shows example image-agnostic perturbations $(\delta)$ crafted by the proposed method. Note that the perturbations look very different for each of the target CNNs. Interestingly, the perturbations corresponding to the VGG models look similar, which might be due to their architectural similarity. Figure~\ref{fig:data-free-sample-images-l2} shows sample perturbed images $(x+\delta)$ for VGG-19~\cite{vgg-arxiv-2014} from ILSVRC~\cite{imagenet-ijcv-2015} validation set. The top row shows the clean and bottom row shows the corresponding adversarial images. Note that the adversarially perturbed images are visually indistinguishable form their corresponding clean images. All the clean images shown in the figure are correctly classified and are successfully fooled by the added perturbation. Below each image, corresponding label predicted by the model is shown. Note that the correct labels are shown in black color and the wrong ones in red.

\subsubsection{Exploiting the minimal prior}
\label{subsub:minimal-prior}
In this section, we present experimental results to demonstrate how our data-free objective can exploit the additional prior information about the target data distribution as discussed in section~\ref{subsec:additional-priors}. Note that we consider two cases: (i) providing the mean and dynamic range of the data samples, denoted as range prior (sec.~\ref{subsubsec:range-prior}), and (ii) utilizing minimal data samples themselves during the optimization, denoted as data prior (~\ref{subsubsec:data-prior}). 
%%%%%%%%%%% Table-2 Fooling rates vs. various priors ILSVRC  start %%%%%%%%%
\begin{table}[ht]
\centering
\caption{Fooling rates for the proposed objective with and without utilizing prior information about the training data. For comparison, we provide the random baseline, existing data-free~\cite{mopuri-bmvc-2017}, and data dependent~\cite{universal-cvpr-2017} objectives.}
\label{tab:fooling-various-priors}
\setlength\tabcolsep{6pt} 
\begin{tabular}{@{}lcccccc@{}}
\toprule
Model     & Baseline & $P_{NP}$ & $P_{RP}$ & $P_{DP}$ & FFF~\cite{mopuri-bmvc-2017} & UAP~\cite{universal-cvpr-2017} \\ \midrule
CaffeNet     & 12.9 & 84.88    & 87.02       & 91.54   & 80.92   & \textbf{93.1}   \\
VGG-F        & 12.62 & 85.96    & 91.81       & 92.64   & 81.59  & \textbf{93.8}   \\
Googlenet    & 10.29 & 58.62    & 71.44       & \textbf{83.54}   & 56.44  & 78.5   \\
VGG-16       & 8.62 & 45.47    & 63.08       & 77.77   & 47.10  & \textbf{77.8}   \\
VGG-19       & 8.40 & 40.68    & 64.67       & 75.51   & 43.62  & \textbf{80.8}   \\
Resnet-152   & 8.99 & 29.78    & 37.3        & 66.68   &  - & \textbf{84.0}   \\ \bottomrule
\end{tabular}
\end{table}
%%%%%%%%%%% Table-2 Fooling rates vs. various priors ILSVRC  end %%%%%%%%%

Table~\ref{tab:fooling-various-priors} shows the fooling rates obtained with and without utilizing the prior information. Note that all the fooling rates are computed for white-box attacking scenario. %To emphasize the effectiveness of the proposed objective, we present fooling rates obtained by a random baseline perturbation. Since our learned perturbation is norm limited by $\xi$, we sample random $\delta$ from $\mathcal{U}[-\xi, \xi]$ and compute the fooling rates. We denote these results in the `Baseline' column of Table~\ref{tab:fooling-various-priors}. 
For comparison, fooling rates obtained by our previous data-free objective~\cite{mopuri-bmvc-2017} and a data dependent objective~\cite{universal-cvpr-2017} are also presented. Important observations to draw from the table are listed below:
\begin{itemize}
\item Utilizing the prior information consistently improves the fooling ability of the crafted perturbations.
\item A simple range prior can boost the fooling rates on an average by an absolute $10\%$, while still being data-free.
\item Although the proposed objective is not designed to utilize the data, feeding the data samples results in an absolute $22\%$ rise in the fooling rates. Due to this increase in performance, for all models (except ResNet-152) our method becomes comparable or even better than UAP~\cite{universal-cvpr-2017}, which is designed especially to utilizes data.%For some of the models (e.g.\ GoogLeNet and VGG-16) our method performs equally (or even outperforms) to an objective~\cite{universal-cvpr-2017}

%\item Our improved objective~(eqn.\ref{eqn:data-free-objective}) and the optimization procedure~(\ref{subsec:improved-optimization}) result in an absolute increase of fooling rate by $22.36\%$ in the case of utilizing data.
\end{itemize}

\subsection{Semantic segmentation}
\label{subsec:segmentation}

In this subsection, we demonstrate the effectiveness of \textit{GD-UAP} objective to craft universal adversarial perturbations for semantic segmentation. We consider four network architectures. The first two architectures are from FCN~\cite{fcn-pami-2017}: \textbf{FCN-Alex}, based on Alexnet~\cite{caffe-arxiv-2014}, and \textbf{FCN-8s-VGG}, based on the 16-layer VGGNet~\cite{vgg-arxiv-2014}. The last two architectures are 16-layer VGGNet based \textbf{DL-VGG}~\cite{deeplab-iclr-2015}, and \textbf{DL-RN101}~\cite{deeplab-resnet-arxiv-2016}, which is a multi-scale architecture based on ResNet-101~\cite{resnet-2015}.

% ######           Fig-4 begins       #####################################
\begin{figure}[b]
\centering
\noindent\begin{minipage}{\linewidth}
  \centering
  \begin{minipage}{.22\textwidth}
  	\centering
    \includegraphics[width=\linewidth]{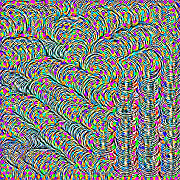}\
    FCN-Alex
  \end{minipage}
   \begin{minipage}{.22\textwidth}
   	\centering
    \includegraphics[width=\linewidth]{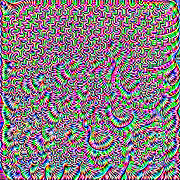}\
    {\small FCN-8s-VGG}
  \end{minipage}
  \begin{minipage}{.22\textwidth}
   \centering
    \includegraphics[width=\linewidth]{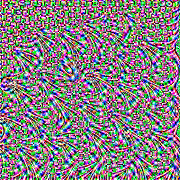}\
    DL-VGG
  \end{minipage}
  \begin{minipage}{.22\textwidth}
  	\centering
    \includegraphics[width=\linewidth]{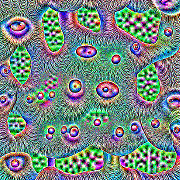}\
    DL-RN101
  \end{minipage}
    \vspace{0.002\textwidth}
\end{minipage}
%\vspace{.01cm}
\caption{Universal adversarial perturbations for semantic segmentation, crafted by the proposed \textit{GD-UAP} objective for multiple models. Perturbations were crafted with ``data w/ less BG'' prior. Images are best viewed in color.}
\label{fig:data-free-sample-perturbations-segmentation}
\end{figure}
The FCN architetures are trained on Pascal VOC-$2011$ dataset~\cite{pascal-voc-2011,hariharan2011semantic}, consisting $9,610$ training samples and the the remaining two architectures are trained on Pascal VOC-$2012$ dataset~\cite{pascal-voc-2012,hariharan2011semantic}, consisting $10,582$ training samples. However, for testing our perturbation's performance, we only use the validation set provided by~\cite{fcn-pami-2017}, which consist of $736$ images.

Semantic segmentation is realized as assigning a label to each of the image pixels. That is, these models are typically trained to perform pixel level classification into one of $21$ categories (including the background) using the cross-entropy loss. Performance is commonly measured in terms of mean IOU (intersection over union) computed between the predicted map and the ground truth. Extending the UAP generation framework provided in~\cite{universal-cvpr-2017} to segmentation is a non-trivial task. However, our generalizable data-free algorithm can be applied for the task of semantic segmentation without any changes.

Similar to recognition setup, we present multiple scenarios for crafting the perturbations ranging from no data to utilizing data samples from the target distribution. An interesting observation with respect to the data samples from Pascal VOC-$2012$, is that, in the $10,582$ training samples, $65.4$\% of the pixels belong to the `background' category. Due to this, when we craft perturbation using training data samples as target distribution prior, our optimization process encounters roughly $65$\% pixels belonging to `background` category, and only $35$\% pixels belonging to the rest $20$ categories. As a result of this data imbalance, the perturbation is not sufficiently capable to corrupt the features of pixels belonging to the categories other than `background'. To handle this issue, we curate a smaller set of $2,833$ training samples from Pascal VOC-$2012$, where each sample has less than $50$\% pixels belonging to `background' category. We denote this as ``data w/ less BG'', and only $33.5$\% of pixels in this dataset belong to the `background' category. The perturbations crafted using this dataset as target distribution prior show a higher capability to corrupt features of pixels belonging to the rest $20$ categories. Since mean IOU is the average of IOU accross the $21$ categories, we further observe that perturbations crafted using ``data w/ less BG" cause a substantial reduction in the mean IOU measure as well.

% ######           Fig-6 begins       #####################################
\begin{figure}[t]
\centering
\noindent\begin{minipage}{\columnwidth}
  \centering
  \begin{minipage}{.18\textwidth}
  	\centering
    \includegraphics[width=\linewidth]{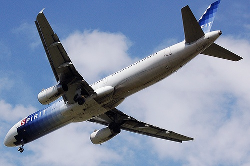}\
  \end{minipage}
   \begin{minipage}{.18\textwidth}
   	\centering
    \includegraphics[width=\linewidth]{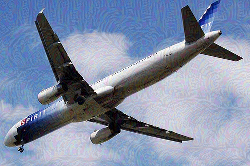}\
  \end{minipage}
  \begin{minipage}{.18\textwidth}
  	\centering
    \includegraphics[width=\linewidth]{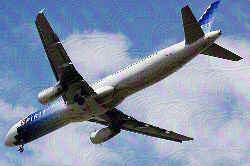}\
  \end{minipage}
  \begin{minipage}{.18\textwidth}
  	\centering
    \includegraphics[width=\linewidth]{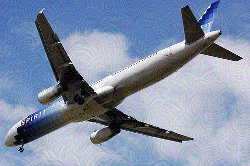}\
  \end{minipage}
  \begin{minipage}{.18\textwidth}
  	\centering
    \includegraphics[width=\linewidth]{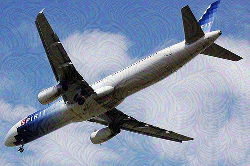}\
  \end{minipage}
  \vspace{0.002\textwidth}
\end{minipage}
\noindent\begin{minipage}{\columnwidth}
  \centering
  \begin{minipage}{.18\textwidth}
  \centering
    \includegraphics[width=\linewidth]{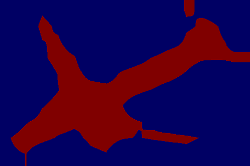}\
    Clean\\Output
  \end{minipage}
   \begin{minipage}{.18\textwidth}
   \centering
    \includegraphics[width=\linewidth]{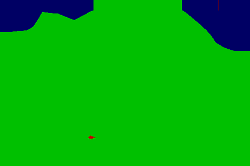}\
    No prior\\Output
  \end{minipage}
   \begin{minipage}{.18\textwidth}
   \centering
    \includegraphics[width=\linewidth]{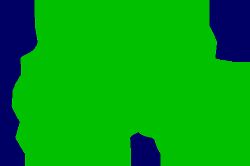}\
    Range \\Prior
  \end{minipage}
  \begin{minipage}{.18\textwidth}
  	\centering
    \includegraphics[width=\linewidth]{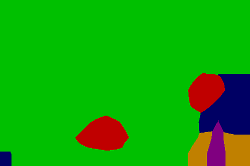}\
    Less BG \\prior
  \end{minipage}
  \begin{minipage}{.18\textwidth}
  \centering
    \includegraphics[width=\linewidth]{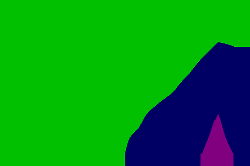}\
    All data \\Prior
  \end{minipage}
    \vspace{0.002\textwidth}
\end{minipage}
%\noindent\begin{minipage}{\columnwidth}
%\centering
%\includegraphics[width=.90\linewidth]{figures/examples/segmentation/fcn_alexnet/2007_006946_color_bar.png}\
%\end{minipage}
%\vspace{0.1cm}
\caption{Sample original and adversarial images from PASCAL-2011 dataset generated for \textbf{FCN-Alex}. First row shows clean and adversarial images with various priors. Second row shows the corresponding predicted segmentation maps.}
\label{fig:sample-segmentation-images}
\end{figure}
% ######           Fig-6 ends         #####################################
% ######           Fig-7 begins       #####################################
\begin{figure}[t]
\centering
\noindent\begin{minipage}{\columnwidth}
  \centering
  \begin{minipage}{.20\linewidth}
  	\centering
    \includegraphics[width=\textwidth]{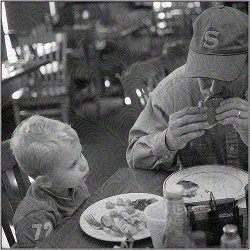}\
  \end{minipage}
   \begin{minipage}{.20\textwidth}
   	\centering
    \includegraphics[width=\textwidth]{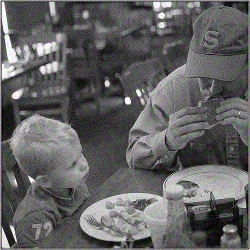}\
  \end{minipage}
  \begin{minipage}{.20\textwidth}
  	\centering
    \includegraphics[width=\textwidth]{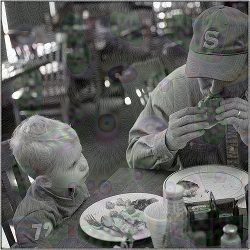}\
  \end{minipage}
  \begin{minipage}{.20\textwidth}
  	\centering
    \includegraphics[width=\textwidth]{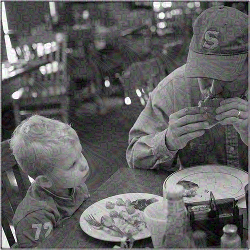}\
  \end{minipage}
  \vspace{0.002\textwidth}
\end{minipage}
\noindent\begin{minipage}{\columnwidth}
  \centering
  \begin{minipage}{.20\textwidth}
  	\centering
    \includegraphics[width=\textwidth]{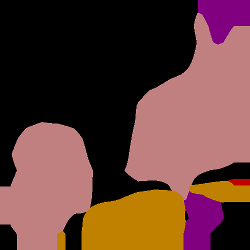}\
  \end{minipage}
   \begin{minipage}{.20\textwidth}
   	\centering
    \includegraphics[width=\textwidth]{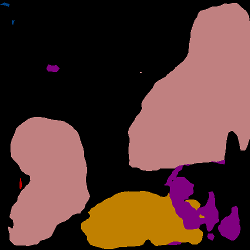}\
  \end{minipage}
  \begin{minipage}{.20\textwidth}
  	\centering
    \includegraphics[width=\textwidth]{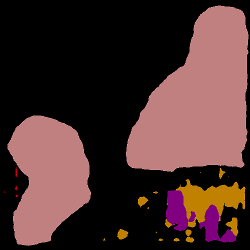}\
  \end{minipage}
  \begin{minipage}{.20\textwidth}
  	\centering
    \includegraphics[width=\textwidth]{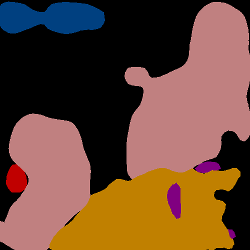}\
  \end{minipage}
  \vspace{0.004\textwidth}
\end{minipage}
\noindent\begin{minipage}{\columnwidth}
  \centering
  \begin{minipage}{.20\textwidth}
  \centering
    \includegraphics[width=\textwidth]{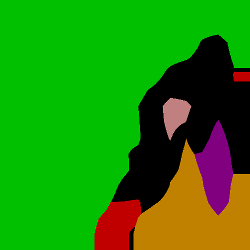}\
    FCN\\Alex
  \end{minipage}
   \begin{minipage}{.2\textwidth}
   \centering
    \includegraphics[width=\textwidth]{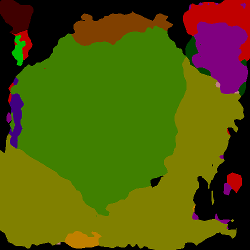}\
    FCN-8S\\VGG
  \end{minipage}
  \begin{minipage}{.2\textwidth}
  	\centering
    \includegraphics[width=\textwidth]{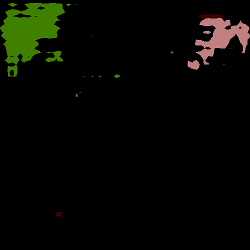}\
    DL\\RN101
  \end{minipage}
  \begin{minipage}{.2\textwidth}
  \centering
    \includegraphics[width=\textwidth]{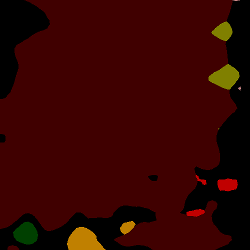}\
    DL\\VGG
  \end{minipage}
    \vspace{0.002\textwidth}
\end{minipage}
\vspace{0.1cm}
%\noindent\begin{minipage}{\columnwidth}
%\centering
%\includegraphics[width=.8\linewidth]{figures/examples/segmentation/all_models/2007_009413_color_bar.png}\
%\end{minipage}
\caption{Segmentation predictions of multiple models over a sample perturbed image. Perturbations were crafted using the ``data w/ less BG'' prior. The first row shows the perturbed input image, the second shows the segmentation output of clean sample image, and the third shows the segmentation output of the perturbed sample image. Images are best viewed in color.}
\label{fig:multiple-nets-segmentation}
\end{figure}
% ######           Fig-7 ends         #####################################
 %%%%%%%%%%% Table-5 GFR for segmentation start %%%%%%%%%
\begin {table}[]
\centering
\caption{Generalized fooling rates achieved by the perturbations crafted by the proposed approach under various settings. Note that for comparison, fooling rates achieved by random perturbations are also presented.}
\label{tab:segmentation-fooling-rates}
\begin{tabular}{lccccc}
\toprule
Model        & Baseline & \pbox{20cm}{No \\ Data} & \pbox{20cm}{Range \\ Prior} & \pbox{20cm}{All \\ Data} & \pbox{20cm}{Data W/ \\ less BG} \\ \midrule
FCN-Alex & 14.29    & 80.15   & 86.57 &85.96    & \textbf{89.61}        \\
FCN-8s-VGG   & 9.24     & 49.42   & 55.04 &61.15    & \textbf{67.19}        \\
DL-VGG   & 10.66    & 55.90 & 58.96 & 44.82    & \textbf{66.68}        \\
DL-RN101  & 8.8      & 37.06   & 35.6 &\textbf{58.62}    & 56.61       \\ \midrule
\end{tabular}
\end{table}
%%%%%%%%%%% Table-5 GFR for segmentation end %%%%%%%%%
Table~\ref{tab:segmentation-fooling-rates} shows the generalized fooling rates with respect to the mean IOU ($GFR(mIOU)$) obtained by \textit{GD-UAP} perturbations under various data priors. As explained in section~\ref{subsec:GFR}, the generalized fooling rate measures the change in the performance of a network with respect to a given metric, which in our case is the mean IOU. Note that, similar to the recognition case, the fooling performance monotonically increases with the addition of data priors. This observation emphasizes that the proposed objective, though being an indirect, can rightly exploit the additional prior information about the training data distribution. Also, for all the models (Other than \textbf{DL-RN101}), ``data w/ less BG'' scenario results in the best fooling rate. This can be attributed to the fact that in ``data w/ less BG'' scenario we reduce the data-imbalance which in turn helps to craft perturbations that fool both background and object pixels.% For comparison, we present the fooling rates achieved by the random perturbation sampled from $\mathcal{U}[-\xi, \xi]$. Similar to recognition experiments, for all the segmentation experiments we use a $\xi$ value of $10$.

In Table~\ref{tab:segmentation-miou} we present the mean IOU metric obtained on the perturbed images learned under various scenarios along with original mean IOU obtained on clean images. %Comparison with baseline random noise perturbation sampled from $\mathcal{U}[-10,10]$ is also provided. 
It is clearly observed that the random perturbation (the baseline) is not effective in fooling the segmentation models. However, the proposed objective crafts perturbations within the same range that can significantly fool the models. We also show the mean IOU obtained by Xie \textit{et al.\ }~\cite{segmentation-detection-iccv-2017}, an image specific adversarial perturbation crafting work. Note that since \textit{GD-UAP} is an image-agnostic approach, it is unfair to expect similar performance as \cite{segmentation-detection-iccv-2017}. Further, the mean IOU shown by~\cite{segmentation-detection-iccv-2017} for \textbf{DL-VGG} and \textbf{DL-RN101} models (bottom $2$ rows of Table~\ref{tab:segmentation-miou}) denote the transfer performance, i.e., black-box attacking and hence show a smaller drop of the mean IOU from that of clean images. However, they are provide as an anchor point for evaluating image-agnostic perturbations generated using \textit{GD-UAP}. %as a means of weak comparison instead of no comparison. %since, there are n anchor point for comparing image-agnostic perturbations generated from our proposed objective.

%%%%%%%%%%% Table-6 MIoU for segmentation start %%%%%%%%%
\begin{table}[t]
\centering
\caption{Comparison of mean IOU obtained by various models against \textit{GD-UAP} perturbations. Comparison with image specific adversaries~\cite{segmentation-detection-iccv-2017} is also presented. $*$ denotes being image-specific and $\dagger$ denotes a transfer attack (black-box attacking). Being image specific, \cite{segmentation-detection-iccv-2017} (ICCV $2017$) outperforms our perturbations, however, even our no data perturbations cause more drop in mIOU than their transfer perturbations. }
\label{tab:segmentation-miou}
\setlength\tabcolsep{2pt} 
\begin{tabular}{lccccccc}
\toprule
Model        & Original & Baseline & \pbox{20cm}{No \\ Data} & \pbox{20cm}{Range \\ prior} & \pbox{20cm}{All \\ data} & \pbox{20cm}{Data w/ \\ less BG} &  \cite{segmentation-detection-iccv-2017} \\ \midrule
FCN-Alex & 46.21    & 45.72    &15.35  & 10.37 & 10.64    & 8.03         & 3.98*      \\
FCN-8s-VGG  & 65.49    & 64.34    &42.78  & 39.08 & 33.61    & 28.05        & 4.02*      \\
DL-VGG   & 62.10    & 61.13   & 36.91 & 35.41  & 44.90    & 27.41        & 43.96$^{*\dagger}$     \\
DL-RN101 & 74.94    & 73.42    & 56.40   & 58.66 & 37.45    & 39.00        & 73.01$^{*\dagger}$    \\ \midrule
\end{tabular}
\end{table}

Figure~\ref{fig:data-free-sample-perturbations-segmentation} %and ~\ref{fig:sample-perturbations-evolution-segmentation} 
shows sample image-agnostic adversarial perturbations learned by our objective for semantic segmentation. In Figure~\ref{fig:data-free-sample-perturbations-segmentation}, we show the perturbations learned with ``data w/ less BG'' prior for all the models. Similar to the recognition case, these perturbations look different across architectures. %In Figure~\ref{fig:sample-perturbations-evolution-segmentation}, we show the perturbations learned for \textbf{FCN-8s-VGG} model under various scenarios ranging from no prior to full data prior. Note that, even for a given network, the  perturbations learned with various priors look different. 
Figures~\ref{fig:sample-segmentation-images} shows example image and predicted segmentation outputs by \textbf{FCN-Alex} model for perturbations crafted with various priors. Top row shows the clean and the perturbed images. Bottom row shows the predictions for the corresponding inputs. Further, the type of prior utilized to craft the perturbation is mentioned below the predictions. Crafted perturbations are clearly successful in misleading the model to predict inaccurate segmentation maps. %Note that the color map shown below the predictions provides the labels. 

Figure~\ref{fig:multiple-nets-segmentation} shows the effect of perturbation on multiple networks. It shows the output maps predicted by various models for the same input perturbed with corresponding $\delta$ learned with ``data w/ less BG" prior. It is interesting to note from Figure~\ref{fig:multiple-nets-segmentation} that for the same image, with UAPs crafted using the same prior, different networks can have very different outputs, even if their outputs for clean images are very similar.

\subsection{Depth estimation}
\label{subsec:depth-estimation}
% ######           Fig-8 begins       #####################################
\begin{figure*}[ht]
\centering
\noindent\begin{minipage}{\textwidth}
  \centering
  \begin{minipage}{.23\textwidth}
  	\centering
    \includegraphics[width=\linewidth]{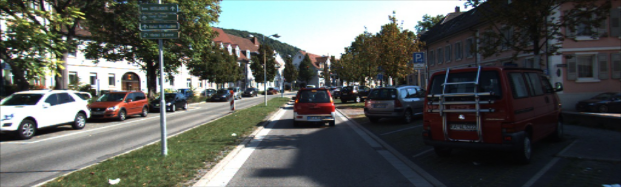}\
  \end{minipage}
   \begin{minipage}{.23\textwidth}
   	\centering
    \includegraphics[width=\linewidth]{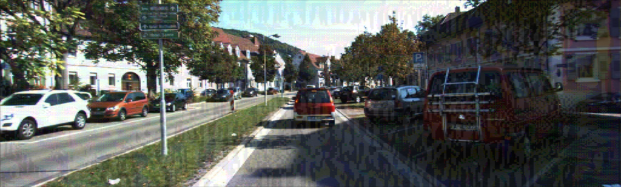}\
  \end{minipage}
  \begin{minipage}{.23\textwidth}
  	\centering
    \includegraphics[width=\linewidth]{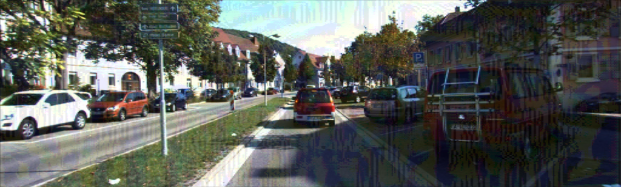}\
  \end{minipage}
  \begin{minipage}{.23\textwidth}
  	\centering
    \includegraphics[width=\linewidth]{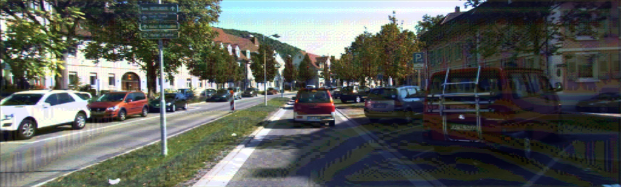}\
  \end{minipage}
  \vspace{0.002\textwidth}
\end{minipage}
\noindent\begin{minipage}{\textwidth}
  \centering
  \begin{minipage}{.23\textwidth}
  	\centering
    \includegraphics[width=\linewidth]{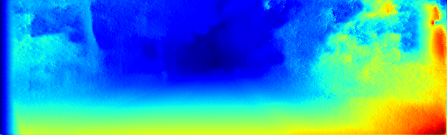}\
  \end{minipage}
   \begin{minipage}{.23\textwidth}
   	\centering
    \includegraphics[width=\linewidth]{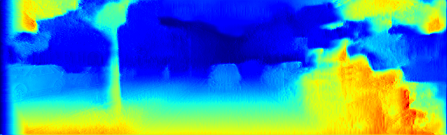}\
  \end{minipage}
  \begin{minipage}{.23\textwidth}
  	\centering
    \includegraphics[width=\linewidth]{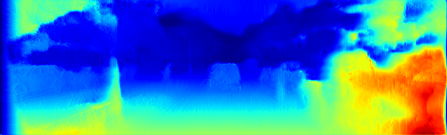}\
  \end{minipage}
  \begin{minipage}{.23\textwidth}
  	\centering
    \includegraphics[width=\linewidth]{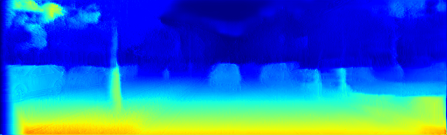}\
  \end{minipage}
  \vspace{0.002\textwidth}
\end{minipage}
\vspace{0.1cm}
\caption{Sample original and adversarial image pairs from KITTI dataset generated for \textbf{Monodepth-VGG}. First row shows clean and perturbed images with various priors. Second row shows the corresponding predicted depth maps.}
\label{fig:sample-depth-images-vgg}
\end{figure*}
% ######           Fig-8 ends         #####################################

Recent works such as~\cite{monodepth-cvpr-2017,8099912,8100177} show an increase in use of convolutional networks for regression-based computer vision task. A natural question to ask is whether they are as susceptible to universal adversarial attacks, as CNNs used for classification. In this section, by crafting UAPs for convolutional networks performing regression, we show that they are equally susceptible to universal adversarial attacks. To the best of our knowledge, we are the first to provide an algorithm for crafting universal adversarial attacks for convolutional networks performing regression, %{\color{blue}though image specific perturbations exist~\cite{cisse2017houdini}}.

Many recent works like~\cite{monodepth-cvpr-2017,8099511,8099508} perform depth estimation using convolutional network. In~\cite{monodepth-cvpr-2017}, the authors introduce Monodepth, an encoder-decoder architecture which regresses the depth of given monocular input image.  We craft UAP using \textit{GD-UAP} algorithm for the two variants of Monodepth, \textbf{Monodepth-VGG} and \textbf{Monodepth-ResNet50}. The network is trained using KITTI dataset~\cite{Geiger2012CVPR}. In its
raw form, the dataset contains $42,382$ rectified stereo pairs from $61$ scenes, with a typical image being $1242 \times 375$ pixels in size. We show results on the eigen split, introduced in~\cite{NIPS2014_5539}, which consist of $23,488$ images for training and validation, and $697$ images for test. We use the same crop size as suggested by the authors of~\cite{NIPS2014_5539} and evaluate at the input image resolution.

As in the case of object recognition, UAPs crafted by the proposed method for monodepth also show the potential to exploit priors about the data distribution. We consider three cases, (i) providing no priors ($P_{NP}$), (ii)range prior ($P_{RP}$), and (ii) data prior ($P_{DP}$). For providing data priors, we randomly pick $10,000$ image samples from the KITTI train dataset. To attain complete independence of target data, we perform validation on a set of $1,000$ randomly picked images from Places-205 dataset. The optimization procedure followed is the same as in the case of the previous two task.

\begin{table}[t]
\centering
\caption{Performance of the crafted perturbations for  \textbf{Monodepth-Resnet50}  and \textbf{Monodepth-VGG} using various metrics for evaluating depth estimation on the eigen test-split of KITTI datset. Results are also presented for the clean data (Normal) and the train set mean. The best fooling results for each scenario are shown in bold. The evaluation of train-set mean performance has been taken from~\cite{monodepth-cvpr-2017}. Note that the first four metrics are error based (higher means better fooling) and the later two are precision based (lower is better fooling).}
\label{tab:segmentation-miou-resnet}
% \resizebox{\linewidth}{!}{%
\setlength\tabcolsep{4pt} 
\begin{tabular}{lcccccc}
\toprule
 \scriptsize Metrics & \scriptsize Abs Rel&\scriptsize  Sq Rel &\scriptsize RMSE  &\scriptsize RMSE log & \scriptsize $\delta(1.25)$  &\scriptsize $\delta(1.25^3)$ \\  \midrule
%& Abs Rel  & Sq Rel & RMSE  & RMSE log & $ \delta < 1.25$  & $ \delta < 1.25^2 $ & $ \delta < 1.25^3$ \\ \midrule 
 & \multicolumn{5}{c}{\textbf{Monodepth-ResNet50}} \\ \midrule
Normal & 0.133 & 1.148 & 5.549 & 0.230 & 0.829 & 0.970 \\
Baseline & 0.1339 & 1.1591 & 5.576 & 0.231 & 0.827 & 0.969 \\
$P_{NP}$ & 0.201 & 1.810 & 6.603 & 0.352 & 0.688 &  0.908 \\
$P_{RP}$ & 0.319 & 3.292 & 9.064 & \textbf{0.640} & \textbf{0.460} & \textbf{0.717} \\
$P_{DP}$ & \textbf{0.380} & \textbf{10.278} & \textbf{10.976} & 0.402 & 0.708 &  0.900 \\
Train-mean & 0.361 & 4.826 & 8.102 & 0.377 & 0.638 &  0.894 \\ \midrule
 & \multicolumn{5}{c}{\textbf{Monodepth-VGG}} \\ \midrule
 
Normal & 0.148 & 1.344 & 5.927 & 0.247 & 0.803 &  0.964 \\
Baseline & 0.149 & 1.353 & 5.949 & 0.248 & 0.800 &  0.963 \\
$P_{NP}$ & 0.192 & 1.802 & 6.626 & 0.325 & 0.704 & 0.929 \\
$P_{RP}$& 0.212 & 2.073 & 6.994 & \textbf{0.364} & \textbf{0.658} &  \textbf{0.906} \\
$P_{DP}$  & \textbf{0.355} & \textbf{9.612} & \textbf{10.592} & 0.390 & 0.714 &  0.908 \\
Train-mean & 0.361 & 4.826 & 8.102 & 0.377 & 0.638 &  0.894 \\ \midrule

\end{tabular}

\end{table}
Table~\ref{tab:segmentation-miou-resnet} %and~\ref{tab:segmentation-miou-vgg} 
show the performance of \textbf{Monodepth-Resnet50} and \textbf{Monodepth-VGG}%, and \textbf{Monodepth-VGG} respectively
 under the presence of the various UAPs crafted by the proposed method. As can be observed from the table, the crafted UAPs have a strong impact on the performance of the network. For both the variants of monodepth, UAPs crafted with range prior, bring down the accuracy with the threshold of $1.25$ units ($ \delta < 1.25 $) by $25.7$\% on an average. With data priors, the crafted UAPs are able to increase the Sq Rel (an error metric) to almost $10$ times the original performance. Under the impact of the crafted UAPs, the network's performance drops below that of the depth-baseline (\textit{Train-mean}), which uses the train set mean as the prediction for all image pixels. Figure~\ref{fig:sample-depth-images-vgg} shows the input-output pair for \textbf{Monodepth-VGG}, where the input is perturbed by the various kinds of UAPs crafted. %Figure~\ref{fig:sample-depth-images-both} compares the input-output pair for \textbf{Monodepth-VGG} and \textbf{Monodepth-Resnet50} using the range-prior UAP. As show by figures~\ref{fig:sample-depth-images-vgg} and~\ref{fig:sample-depth-images-both}, for both versions of Monodepth, quasi-imperceptible change in the input can cause a large change in the output. 

Table~\ref{tab:fooling-rate-depth} shows the Generalized Fooling Rates (GFR) with respect $ \delta < 1.25 $, i.e. $GFR(\delta < 1.25)$. It is observed that $P_{RP}$ has higher $GFR(\delta < 1.25)$ than $P_{DP}$. This may appear as an anomaly as $P_{DP}$, which has access to more information, should cause stronger harm to the network than $P_{RP}$. This is indeed reflected in terms of multiple metrics such as $Abs. Rel. Error$, and $RMSE$ (ref. Table~\ref{tab:segmentation-miou-resnet}). In fact, $P_{DP}$ is able to reduce these metrics even below the values achieved by the train-set mean. This clearly shows that $P_{DP}$ is indeed stronger than $P_{RP}$ (in terms of these metrics).

However, in terms of the other metrics, such as $\delta <1.25$ and $\delta <1.25^3$ , it is observed that $P_{RP}$ causes more harm. These metrics measure the $\%$ of pixels where $f(x) - G(x)$ (where $G(x)$ represents the ground truth depth at $x$) is lesser than pre-defined limits. In contrast, metrics such as  $Abs. Rel. Error$, and $RMSE$ measure the overall error of the output. Hence, based on the effect of $P_{DP}$ and $P_{RP}$ on these metrics, we infer that while $P_{DP}$ shifts fewer pixels than $P_{RP}$, it severely shifts those pixels. In fact, as shown in Figure~\ref{fig:sample-depth-images-vgg}, it is often noticed that $P_{DP}$ causes the network to completely miss some nearby objects, and hence predicting very high depth at such locations, whereas $P_{RP}$ causes a anomalous estimation at a higher number of pixels.

The above situation shows that the `fooling' performance of a perturbation can vary based on the metric used for analysis. Further, conclusions based on a single metric may only partially reflect the truth. This motivated us to propose $GFR(m)$, a metric dependant measurement of `fooling', which clearly indicates the metric dependence of `fooling'.%This also highlights that, `which perturbation is better?', is rather a metric dependant question and conclusions based on a single metric would only partially reflect the truth.

%Table~\ref{tab:fooling-rate-depth} shows the Generalized Fooling Rates (GFR) with respect to the accuracy with the threshold of $1.25$ units ($ \delta < 1.25 $). A surprising observation from the table is that the performance of the range prior perturbations on both the networks surpasses that of the data prior perturbations. This might lead us to understand that the range prior perturbations are stronger. However the Generalized Fooling Rate (GFR) measures the change in a network's outcome, with respect to a metric. As observed from Table~\ref{tab:segmentation-miou-resnet} and~\ref{tab:segmentation-miou-vgg}, the data prior perturbations have a much harsher effect on the error metrics like Root Mean Square Error (RMSE), whereas range prior perturbations have a harsher effect on precision metrics like $ \delta < 1.25 $. Hence, `which perturbation is better?', is rather a metric dependant question and conclusions based on a single metric would only partially reflect the truth.

%Figures~\ref{fig:sample-depth-images-vgg} and~\ref{fig:sample-depth-images-both} show some qualitative examples, showing the perturbed input and the change in output.

%%%%%%%%%%% Table-8 metrics for depth estimation using Monodepth-VGG end %%%%%%%%%
\begin{table}[]
\centering
\caption{GFR with respect to $ \delta < 1.25$ metric for the task of depth estimation}
\label{tab:fooling-rate-depth}
\begin{tabular}{lcccc}
\toprule
Model        & Baseline & \pbox{20cm}{No \\ data} & \pbox{20cm}{Range \\ prior}  & \pbox{20cm}{Data \\ prior} \\ \midrule 
Monodepth-VGG & 0.4\% &  15.3\% & \textbf{22.7\%} & 21.3\% \\
Monodepth-Resnet50 & 2\% & 21.3\% & \textbf{47.6\%} & 24.3\%  \\\midrule
\end{tabular}
\end{table}

\section{\textit{GD-UAP}: Analysis and Discussion}
\label{sec:anal}
In this section, we provide additional analysis of \textit{GD-UAP} on various fronts. First, we clearly highlight the multiple advantages of \textit{GD-UAP} by comparing it with other approaches. In the next subsection we provide a thorough experimental evaluation of image-agnostic perturbation in the presence of various defence mechanism. Finally, we end this section with a discussion on how \textit{GD-UAP} perturbations work.

\subsection{Comparison with other approaches}
\subsubsection{Comparison of data-free objective}
\label{subsec:objectives-comparison}
%%%%%%%%%%% Table-3 mean activation vs. l2 norm  start %%%%%%%%%
\begin{table}[t]
\centering
\caption{Comparison of data-free objectives. Fooling rates achieved by maximizing $l_2$ norm (\textit{GD-UAP}) vs. mean activation (~\cite{mopuri-bmvc-2017}) when utilizing data samples. %Note that the proposed objective consistently outperforms the previous objective across multiple architectures (shown in bold).
}
\label{tab:objectives-comparison}
\begin{tabular}{@{}lcc@{}}
\toprule
Model      & FFF~\cite{mopuri-bmvc-2017} & $l_2$ GD-UAP  \\ \midrule
CaffeNet   & 88.35          & \textbf{91.54}          \\
VGG-16     & 72.68          & \textbf{77.77}           \\
Resnet-152 & 65.43          & \textbf{66.68}          \\ \bottomrule
\end{tabular}
\end{table}
%%%%%%%%%%% Table-3 mean activation vs. l2 norm  end %%%%%%%%%
First, we compare the effectiveness of \textit{GD-UAP} against the existing data-free objective proposed in~\cite{mopuri-bmvc-2017}. Specifically, we compare maximizing the mean versus $l_2$ norm (energy) of the activations caused by the perturbation $\delta$ (or $x/d+\delta$ in case of exploiting the additional priors).

Table~\ref{tab:objectives-comparison} shows the comparison of fooling rates obtained with both the objectives (separately) in the improved optimization setup (\ref{subsec:improved-optimization}). We have chosen $3$ representative models across various generations of models (CaffeNet, VGG and ResNet) to compare the effectiveness of the proposed objective. Note that the improved objective consistently outperforms the previous one by a significant $3.18\%$. Similar behaviour is observed for other vision tasks also. %This shows that maximizing the activations caused by the perturbation can lead to 
\subsubsection{Data dependent vs. Data-free objectives}
\label{subsubsec:dd-vs-df}
%%%%%%%%%%% Table-4 effect of data dependency  start %%%%%%%%%
\begin{table}
\centering
\caption{Effect of data dependency on crafting the perturbations. Data dependent objectives~\cite{universal-cvpr-2017} suffer significant drop in fooling ability when arbitrary data samples are utilized for crafting. A$\rightarrow$ B denotes that data A is used to craft perturbations to fool models trained on data B. Note that fooling rates for our approach are crafted without utilizing any data samples (denoted with $^*$).}
\vspace{-0.2cm}
\label{tab:dd-vs-df}
\begin{tabular}{lcccc}
\toprule
Model     & \multicolumn{2}{c}{Places-205 $ \rightarrow $ ILSVRC} & \multicolumn{2}{c}{ILSVRC $\rightarrow$ Places-$205$} \\ \midrule
          & Ours   & UAP~\cite{universal-cvpr-2017}  & Ours   & UAP~\cite{universal-cvpr-2017}  \\ \midrule
CaffeNet  & \textbf{87.02}*  & 73.09                                  & \textbf{88.61}*  & 77.21                                  \\
GoogLenet & \textbf{71.44}*  & 28.17                                  & \textbf{83.37}*  & 52.53 \\ \bottomrule                                
\end{tabular}
\end{table}
%%%%%%%%%%% Table-4 effect of data dependency  end %%%%%%%%%
Now, we demonstrate the necessity of data dependent objective~\cite{universal-cvpr-2017} to have samples from the target distribution only. That is, methods (such as~\cite{universal-cvpr-2017}) that craft perturbations with fooling objective (i.e.\ move samples across the classification boundaries) require samples from only the training data distribution during the optimization. We show that crafting with arbitrary data samples leads to significantly inferior fooling performance.

Table~\ref{tab:dd-vs-df} shows the fooling rates of data dependent objective~\cite{universal-cvpr-2017} when non-target data samples are utilized in place of target samples. Experiment in which we use Places-$205$ data to craft perturbations for models trained on ILSVRC is denoted as Places-$205$ $\rightarrow$ ILSVRC and vice versa. For both the setups, a set of $10,000$ training images are used. Note that, the rates for the proposed method are obtained without utilizing any data (with range prior) and rates for data-free scenario can be found in Table~\ref{tab:fooling-various-priors}. Clearly the fooling rates for UAP~\cite{universal-cvpr-2017} suffer significantly, as their perturbations are strongly tied to the target data. On the other hand, for the proposed method, since it does not craft via optimizing a fooling objective, the fooling performance does not decrease. Importantly, these experiments show that the data dependent objectives are not effective when samples from the target distribution are not available. This is a major drawback as it is difficult to procure the training data in practical scenarios. 
\begin{figure}[t]
    \centering
    \includegraphics[width=\linewidth]{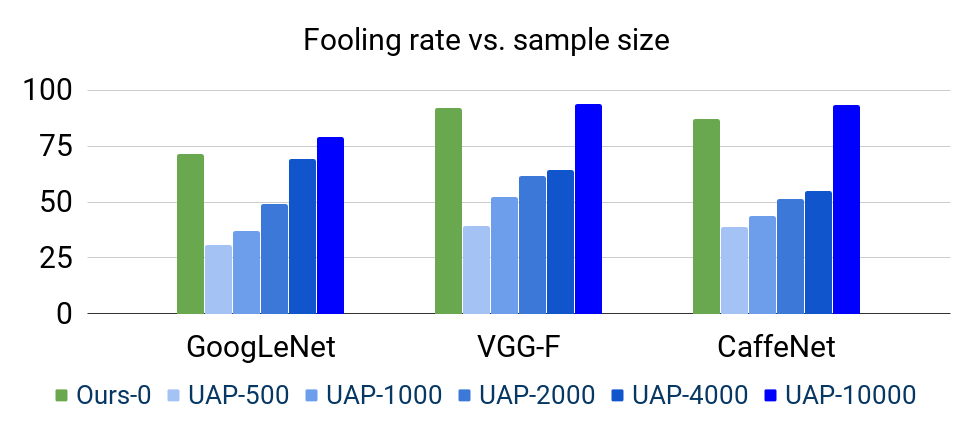}
    \caption{Reliance of the data dependent objective UAP~\cite{universal-cvpr-2017} on the size of available training data samples. Note that our approach utilizes no data samples and achieves competitive fooling performance.}
    \label{fig:dd-vs-df}
\end{figure}
%%%%%%%%%%% Table-6 MIoU for segmentation end %%%%%%%%%

Additionally, as the data dependent objectives rely on the available training data, the ability of the crafted perturbations heavily depends on the size of the available training data. We show that the fooling performance of UAP~\cite{universal-cvpr-2017} significantly decreases as the size of the available samples decreases. Figure~\ref{fig:dd-vs-df} shows the fooling rates obtained by the perturbations crafted for multiple recognition models trained on ILSVRC by UAP~\cite{universal-cvpr-2017} with varying size of samples available for optimization. We craft different UAP~\cite{universal-cvpr-2017} perturbations (using the codes provided by the authors) utilizing only $500$, $1,000$, $2,000$, $4,000$ and $10,000$ data samples and evaluate their ability to fool various models. The performance of the crafted perturbations decreases drastically (shown in different shades of blue) as the available data samples are decreased during the optimization. For comparison, fooling rates obtained by the proposed data-free objective is shown in green.

\begin{table}[]
\centering
\caption{We present the fooling rate comparison of the existing data dependent UAP[8] approach and the  proposed approach. }
\label{tab:fair-comparison}
\begin{tabular}{lll}
\toprule
Priors       & Ours  & UAP{[}8{]}     \\ \midrule
No data     & 58.62 & No convergence \\ 
Range Prior & 71.44 & 10.56         \\ 
Data Prior  & 83.54 & 78.5              \\ 
\bottomrule
\end{tabular}
\end{table}

\subsubsection{Capacity to expoit minimal priors}
\label{subsubsec:dd-vs-df}
An interesting question to ask is whether the data dependant approach UAP~\cite{universal-cvpr-2017} can also utilize minimal data priors. To answer this, we craft perturbations using the algorithm presented in UAP, with different data priors (including no data priors). Table~\ref{tab:fair-comparison} presents the comparison of \textit{GD-UAP} against UAP~\cite{universal-cvpr-2017} with different priors. The numbers are computed on the ILSVRC validation set for Googlenet. We observe that the UAP algorithm does not even converge in the absence of data, failing to create a data-free UAP. When range prior (Gaussian noise in the data range) is used to craft, the resulting perturbations demonstrate a very low fooling rate of $10.56$ compared to $71.44$ of the proposed method. This is not surprising, as a significant decrease in the performance of UAP can be observed even due to a simple mismatch of training and target data (ref Table~\ref{tab:dd-vs-df}). Finally, when actual data samples are used for crafting, UAP~\cite{universal-cvpr-2017} achieves $78.5$ success rate which is closer to $83.54$ of the proposed method.

From these results, we infer that since UAP~\cite{universal-cvpr-2017} is a data dependent method, it requires corresponding data samples to craft effective perturbations. When noise samples are presented for optimization, the resulting perturbations fail to generalize to the actual data samples. In contrast, \textit{GD-UAP} solves for an activation objective and exploits the prior as available.

\subsection{Robustness of UAPs Against Defense Mechanisms}
While many recent works propose novel approaches to craft UAPs, the current literature does not contain a thorough analysis of UAPs in the presence of various defense mechanisms. If simple defense techniques could render them harmless, they may not present severe threat to deployment of deep models. In this subsection, we investigate the strength of \textit{GD-UAP} against various defense techniques. Particularly, we consider (1) \textit{Input Transformation Defenses}, such as Gaussian blurring, Image quilting, etc., (2) \textit{Targeted Defense Against UAPs}, such as Perturbation Rectifying Networks (PRN)~\cite{defence-uap-arxiv2017} and (3) \textit{Defense through robust architectural design} such as scattering networks~\cite{hybrid-net}.

%While many recent approaches have been proposed for crafting Universal Adversarial Perturbations (\cite{universal-cvpr-2017, universalseg-iccv-2017}), they do not present rigorous study of their robustness against various defense techniques. If simple defense techniques could render them harmless, they may not present severe threat to deployment of deep models. This leads us to conduct extensive experimentation about the ability of the crafted perturbations against defenses. 

We evaluate the performance of UAPs generated from \textit{GD-UAP}, as well as data-dependent approach UAP~\cite{universal-cvpr-2017} against various defense mechanisms. %We divide the defenses into two kinds: (1) \textbf{Defense by input transformations}, and (2) \textbf{Targeted Defense for UAPs}. 
\begin{table}[h]
\centering
\caption{The fooling rates computed for various UAP algorithms for under different defenses on GoogLeNet. $Acc_{top1}$ represents the Top-1 Accuracy on clean images in the presence of defense mechanism.}
\label{tab:defense_input}
\begin{tabular}{cccccc}
\toprule
\multicolumn{1}{c}{\multirow{2}{*}{Model}} &\multicolumn{1}{c}{\multirow{2}{*}{$Acc_{top1}$}} &  \multicolumn{2}{c}{Data-Independant} & \multicolumn{2}{c}{Data-Dependant} \\ \cline{3-6} 
\multicolumn{1}{c}{}& \multicolumn{1}{c}{}&  $P_{NP}$       & $P_{RP}$       & $P_{DP}$ & UAP \cite{universal-cvpr-2017}      \\ \midrule
No Defense & 69.74 & 58.62        & 71.44       & \textbf{83.54}      & 78.5   \\ \midrule 
\multicolumn{6}{c}{\textbf{Input transformation defenses}} \\  \midrule
10-Crop & 70.94  & 46.6        & 54.3       & \textbf{66.6}      & -           \\\midrule
Gaussian\\ Smoothing & 58.12    & 25.80         & 35.62        & 32.66      & \textbf{32.78} \\
Median\\ Smoothing   & 51.78  & 35.08        & \textbf{46.98}        & 43.96      & 37.02 \\
Bilateral\\ Smoothing  & 60.28  & 17.06        & 21.50        & 22.34      & \textbf{22.74} \\ \midrule
Bit-Reduction\\(3-bit)  & 63.20  & 48.66        & 61.34        & \textbf{72.30}      & 69.62 \\ 
Bit-Reduction\\(2-bit)  & 45.50  & 46.74        & 56.76        & 60.50     & {65.40} \\ \midrule
JPEG (75\%)  & 67.50  &35.66       & 51.22        & \textbf{61.38}     & 42.62\\
JPEG (50\%) & 64.16  & 29.34        & 41.84        &\textbf{44.62}     & 31.40 \\\midrule 
TV\\ Minimization  & 50.50  & 27.9        & 31.9        & \textbf{30.9}     & 26.22 \\
Image\\ Quilting  & 51.30 & 30.76       & \textbf{36.12}        & 34.80      & 25.84 \\  \midrule
\multicolumn{6}{c}{\textbf{Targeted Defenses Against UAPs}} \\ \midrule
PRN Defense &68.90 & 31.34        & 46.60      & \textbf{52.50}     &21.35           \\ \bottomrule
\end{tabular}
\end{table}

\subsubsection{Input Transformation Defenses}
\label{subsec:defense-input-transformations}
For defense by input transformations, inline with \cite{def_bit} and~\cite{def_jpeg}, we consider the following simple defenses: (1) 10-Crop Evaluation, (2) Gaussian Blurring, (3) Median Smoothing, (4) Bilateral Filtering (5) JPEG Compression, and (6) Bit-Depth Reduction. Further, we evaluate two sophisticated image transformations proposed in~\cite{def_quilt}, namely, (7) TV-Minimization and (8) Image Quilting (Using code provided by the authors).

Table~\ref{tab:defense_input} presents our experimental evaluation of the defenses on the Googlenet. As we can see, while the fooling rate of all UAPs is reduced by the defenses (significantly in some cases), it is achieved at the cost of model's accuracy on clean images. If Image Quilting, or TV-normalization is used as a defense, it is essential to train the network on quilted images, without which, a severe drop in accuracy is observed. However, in majority of the defenses, UAPs flip labels for more than  $45\%$ of the images, which indicates threat to deployment. 
Further, as \textit{Bilateral Filtering} and \textit{JPEG compression} show strong defense capability at low cost to accuracy, we evaluate the performance of our GD-UAP perturbations (with range prior) on $6$ classification networks in the presence of these two defenses.
%Further, we evaluate the performance of our GD-UAP perturbations (with range prior) on $6$ classification networks, in the presence of Bilateral Filtering and JPEG compression as defenses. 
 This is shown in Table~\ref{tab:defense_input_all}. 
 We note that when the defense mechanism significantly lowers the fooling rates, a huge price is paid in terms of \% drop in Top-1 Accuracy $(D_{Acc})$, which is unacceptable. This further indicates the poor fit of input transformations as a viable defense.

\begin{table*}[h]
\centering
\caption{The effect of strong input defenses on perturbations crafted from our objective with range-prior. $D_{Acc}$ represents the percentage drop in Top-1 Accuracy on clean images. }
\label{tab:defense_input_all}
\begin{tabular}{ccccccccccccc}
\toprule
\multicolumn{1}{c}{\multirow{2}{*}{Model}} &\multicolumn{2}{c}{Caffenet} &  \multicolumn{2}{c}{VGG-F} & \multicolumn{2}{c}{GoogLeNet}  & \multicolumn{2}{c}{VGG-16} & \multicolumn{2}{c}{VGG-19} & \multicolumn{2}{c}{ResNet-152}\\ \cline{2-13} 
\multicolumn{1}{c}{}& $\% D_{Acc} $& $FR$ & $\% D_{Acc} $& $FR$ & $\% D_{Acc} $& $FR$ & $\% D_{Acc} $& $FR$ & $\% D_{Acc} $& $FR$ & $\% D_{Acc} $& $FR$ \\ \midrule
No Defense  & -  & 87.1       & -        & 91.8    & - &  71.4 & -        & 63.1 & -        & 64.7 &-        & 37.3\\ 
JPEG(50\%)  & 5.23\%  & 72.1       & 0.5\%      &77.5    & 8\% & 41.8& 2.6\%      & 49.2 & 2.4\%        & 64.7 & 4.7\%    & 37.3\\ 
Bilateral  &  15.2\%   &  34.8      & 18.2\%     & 34.8    & 13.5\% & 21.5 & 14.7\%        & 35.8 & 14.2\%        & 28.2&  9.7\%    & 25.4\\ \bottomrule
\end{tabular}
\end{table*}

%\textbf{Details of the Input Transformation Defenses}: For (1) 10-crop predictions, 4 crops (of size $2/3$ of the image) are taken from each corner of the image, along with a center crop. Now the image is horizontally flipped to obtain 5 more crops in similar fashion. For (2) Gaussian Blurring, we select a kernel size of $(5,5)$, and sigma parameter equal to $1.1$ .  (3) Median Smoothing, is performed with a kernel size of $(5,5)$. For (4) Bilateral Filtering, the diameter, Color-dimension sigma and Space-dimension sigma are set to be $(9,75,75)$ respectively. The previous three methods are implemented using OpenCV library. (5) JPEG Compression is performed using the PIL library in python. For (6) Bit-Depth Reduction, we follow the procedure outlined in \cite{def_bit}. For (7) TV-Minimization, the same set of parameters as released by the authors in \cite{def_quilt} is used. Finally, for (8) Image Quilting, we create $100,000$ patches of size $(5,5)$ from $10,000$ images from Imagenet training-set using the code provided by the authors.

\subsubsection{Targeted Defense Against UAPs}
\label{subsec:defense-targeted}
Now, we turn towards defenses which have been specifically engineered for UAPs. We evaluate the performance of the various UAPs, with the Perturbation Rectifying Network(PRN)~\cite{PRN} as a defense. PRN is trained using multiple UAPs generated from the algorithm presented in UAP~\cite{universal-cvpr-2017} to rectify perturbed images. In Table~\ref{tab:defense_input}, we present the fooling rates obtained on Googlenet using various perturbations, using the codes provided by the authors of~\cite{PRN}. While PRN is able to defend against~\cite{universal-cvpr-2017}, it shows very poor performance against our data prior perturbation. This is due to the fact that PRN is trained using UAPs generated from UAP~\cite{universal-cvpr-2017} only, indicating that PRN lacks generalizability to input-agnostic perturbations generated from other approaches. Furthermore, it is also observed that various simple input transformation defenses outperform PRN. Hence, while PRN is an important step towards defenses against UAPs, in its current form, it provides scarce security against UAPs from methods it is not trained on.
\iffalse
\begin{table}[h]
\centering
\caption{The fooling rates computed for various UAP algorithms, on the GoogLeNet network in the presence of Defense Specifically engineered for UAPs. In specific, we show results with PRN \cite{PRN} as a defense mechanism. }
\label{tab:defense_input_prn}
\begin{tabular}{|l|c|c|c|c|}
\hline
\multicolumn{1}{|c|}{\multirow{2}{*}{Model}} &  \multicolumn{2}{c|}{Data-Independant} & \multicolumn{2}{c|}{Data-Dependant} \\ \cline{2-5} 
\multicolumn{1}{|c|}{}&  \pbox{20cm}{Ours\\ No-prior}       & \pbox{20cm}{Ours\\ Range Prior}       & \pbox{20cm}{Ours\\ Data Prior} & UAP \cite{universal-cvpr-2017}      \\ \hline
No Defense &  58.62        & 71.44       & 83.54      & 78.5             \\ \hline
PRN Defense &  35.36        & 51.18      & 57.42     & 37.6            \\ \hline
\end{tabular}
\end{table}
\fi
\subsubsection{Defense through robust architectural design}
\label{subsec:defense-robust-models}

%We further investigate the models which have small Lipschitz-constant for the constituting layers, such as scattering-networks~\cite{scatter_1}. We conduct experiments and find interesting results.

As our optimization process relies on maximizing $|| f(x + \delta)||_2$ by increasing $\Pi || l_i(x + \delta)||_2 \forall i \in \{1,2,...\}$ (where $l_i(.)$ represents the input to layer $l_{i+1}$), one defense against our attack can be to train the network such that the change in the output to a layer $l_i$ minimally effects the output of the network $f(.)$. %That is, $|| f(x + \delta)-f(x)||_2$ is not drastically increased due to increase in $||l_i(x + \delta)-l_i(x)||_2$. 
One method for achieving this target can be to minimize the Lipschitz Constant $K_i$ of each linear transformation in the network. 
As $K_i$ controls the upper bound of the value of $|| f(x + \delta)-f(x)||_2$ with respect to $||l_i(x + \delta)-l_i(x)||_2$, minimizing $K_i \forall i \in \{ 1,2,...\}$ can lead to a stable system where minor variation in input layer do not translate to high variation in output. This would translate to low fooling rate when attacked by adversarial perturbations. % Our goal is now to evaluate whether a network trained with special constraints to achieve low Lipshitz constant for its layers is also more robust to adversarial attacks from perturbations crafted using our proposed objectives.

In ~\cite{scatter_1}, Bruna \textit{et al.} introduce scattering network. This network consist of scattering transform, which linearize the output deformation with respect to small input deformation, ensuring that the Lipschitz constant is $\leq 1$. In \cite{hybrid-net}, a hybrid approach was proposed, which uses scattering transforms in the initial layers, and learn convolutional layers (Res-blocks, in specific) on the transformed output of the scattering layers, making scattering transform based approaches feasible for Imagenet. The proposed Hybrid-network gave performance comparable to ResNet-18 and VGG-13 while containing much lesser layers.

We now evaluate the fooling rate that \textit{GD-UAP} perturbations achieve in the Hybrid-Network, and compare it to fooling rate achieved on VGG-13 and ResNet-18 networks. In the Hybrid-network, ideally we would like to maximize  $||l_i(x + \delta)||_2$ at each of the Res-Block output. However, as $\partial S(x) / \partial x$, where $S(.)$ represents the Scattering Transform, is non-trivial, we only perform black-box attacks on all the networks.

Table~\ref{tab:def-scatter}  shows the results of the black-box attack on Hybrid Networks. Though Hybrid networks on an average decrease the fooling rate by $13\%$ when compared
to other models, they still remain vulnerable. % to image-agnostic perturbations.% A fully scattering-transform based approach may provide more robustness against UAPs.% This verifies that our attack must be reliant on the Lipschitz constant of the network. 

\begin{table}[t]
\centering
\caption{Fooling Rate for Hybrid-network~\cite{hybrid-net}, ResNet-18 and VGG-13 by black-box attack.}
\label{tab:def-scatter}
\begin{tabular}{cccc}
\toprule
 &  No-Prior & Range-Prior & Data-Prior \\ \midrule
&\multicolumn{3}{c}{\textbf{Perturbation from VGG-13}}\\ \midrule
Hybrid-network &  \textbf{28.91}& \textbf{30.96}& \textbf{33.90}\\ 
ResNet-18  &35.08 &39.14& 43.92\\ \midrule 
&\multicolumn{3}{c}{\textbf{Perturbation from ResNet-18}}\\ \midrule
Hybrid-network & \textbf{ 25.30}& \textbf{27.82}&\textbf{39.36} \\ 
VGG-13 &37.82 &44.70& 60.86\\ \bottomrule
\end{tabular}
\end{table}

\subsection{Analyzing how \textit{GD-UAP} works}
\label{sec:anal_uap}
% ######           Fig-11 starts         #####################################
\begin{figure}[t]
    \centering
    \includegraphics[width=0.8\linewidth]{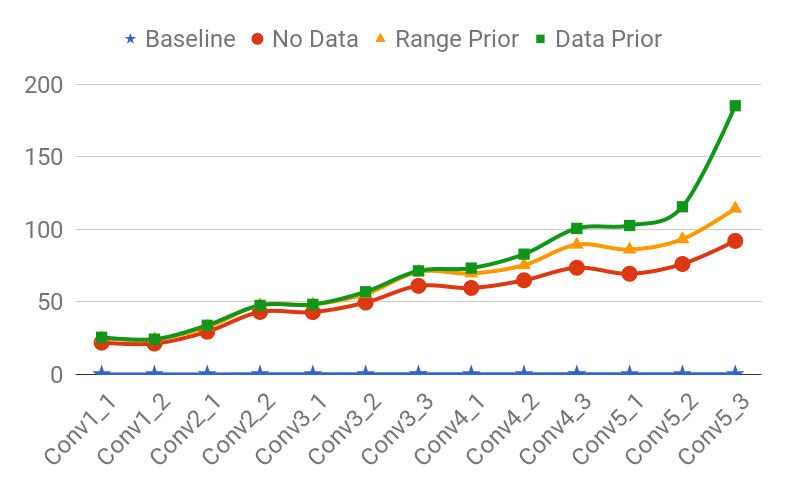}
    \caption{Percentage relative change in the extracted representations caused by our crafted perturbations at multiple layers of VGG-16. %Note that as we do deeper into the net, the amount of perturbation increases drastically which results in eventual misclassification of the input.
    }
    \label{fig:relative-perturbation}

\end{figure}
% ######           Fig-11 ends         #####################################
As demonstrated by our experimentation in section~\ref{sec:expts}, it is evident that \textit{GD-UAP} is able to craft highly effective perturbations for a variety of computer vision tasks. This highlights an important question about the Convolutional Neural Networks (CNNs):``How stable are the learned representations at each layer?" That is, Are the features learned by the CNNs robust to small changes in the input? As mentioned earlier `fooling' refers to instability of the CNN in terms of its output, independent of the task at hand.

%Earlier works~(e.g.,\cite{intriguing-arxiv-2013,explainingharnessing-arxiv-2014,universal-cvpr-2017}) have proven that the CNNs can be fooled by small but learned noise. Most of these works (e.g.,~\cite{explainingharnessing-arxiv-2014}) rely on data to find these directions in which the input can be slightly perturbed in order to fool the CNN. These approaches try to find via simple methods such as gradient ascent, the directions in which the network's confidence falls. Recent works by Moosavi-Dezfooli \textit{et al.}~\cite{deepfool-cvpr-2016,universal-cvpr-2017} find the nearest classification boundary for a given sample and perturb it by moving across the boundary thereby achieve the fooling. They also establish that it is possible to learn a single perturbation which can work for most of the input images. Thus existing works (independent of the underlying task) exploit data samples, model's confidence and the classification boundaries in the input space in order to craft the adversarial perturbations, either image specific or agnostic. 
We depict this as a stability issue with the learned representations by CNNs. We attempt to learn the optimal perturbation in the input space that can cause maximal change in the output of the network. We achieve this via learning perturbations that can result in maximal change in the activations at all the intermediate layers of the architecture. As an example, we consider VGG-$16$ CNN trained for object recognition to illustrate the working of our objective. Figure~\ref{fig:relative-perturbation} shows the percentage relative change in the feature activations $(\frac {\Vert l_i(x+\delta) - l_i(x) \Vert_2 \times 100} {\Vert l_i(x) \Vert_2})$ at various layers in the architecture. The percentage relative change in the feature activations due to the addition of the learned perturbation increases monotonically as we go deeper in the network. Because of this accumulated perturbation in the projection of the input, our learned adversaries are able to fool the CNNs independent of the task at hand. This phenomenon explains the fooling achieved by our objective. We can also observe that with the utilization of the data priors, the relative perturbation further increases which results in better fooling when the prior information is provided during the learning.% Note that, for comparison, the baseline perturbation of random noise with equal max-norm as the learned perturbations is provided. The perturbation caused by the random noise is almost negligible at all the layers of CNN, which explains its robustness to random noise attacks.

\begin{figure}[h]
\centering
\includegraphics[width=0.75\columnwidth]{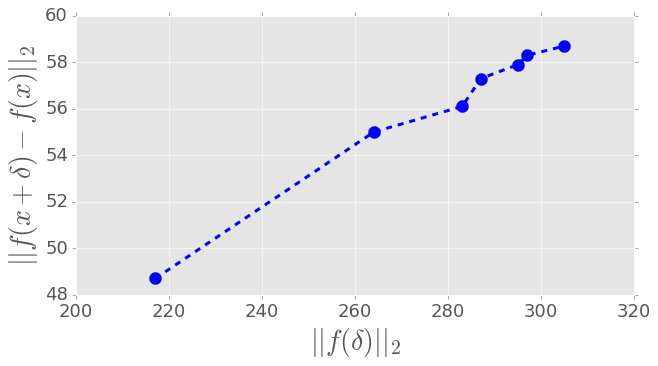}
\caption{Correlation between $|| f(x + \delta) - f(x)||_2$ and $||f(\delta)||_2$ computed for VGG-16 model. Plot shows the fit between $|| f(x + \delta) - f(x)||_2$ and $||f(\delta)||_2$ computed during the training at iterations just before the $\delta$ gets saturated. %Note that the green dotted line indicates the $y=x$ line whereas the blue one is the fit between $|| f(x + \delta) - f(x)||_2$ and $||f(\delta)||_2$ computed during the training at iterations just before the $\delta$ gets saturated and projected back into the constrained set. Also the later portion of the fit is zoomed in for clarity.
}
\label{fig:corrresnet}
\end{figure}

%Recent works such as \cite{}, have shown evidence for the arguement that CNNs are be locally linear. 
Our data-free approach, consists of increasing $||l_i(\delta)||$ across the layers, to increase $f(\delta)$. As Figure~\ref{fig:relative-perturbation} shows, this crafts a perturbation, which leads to an increase in $f(x+\delta)$. One may intrepret this increase to be caused due to the locally linear nature of CNNs. i.e., $f(x + \delta) \approx f(x) + f(\delta)$. However, our experiments reveal that the feature extractor $f$ might not be locally linear. We observe the relation between the quantities $|| f(x + \delta) - f(x)||_2$ and $||f(\delta)||_2$, where $||.||_2$ represents the $L_2$-norm, and $f(.)$ represents the output of the last convolution layer of the network. Figure~\ref{fig:corrresnet} presents the comparison of these two quantities for VGG-16 computed during the proposed optimization with no prior. % The values computed are averaged over $5000$ random class-balanced images from Imagenet validation data. The perturbations are extracted before each rescale operation (projection into the $l_{\infinity}$ ball via division by $2$) during the optimization. 

From the observations, we infer: (1) $|| f(x + \delta) - f(x)||_2$ is \textbf{not} approximately equal to $||f(\delta)||_2$, and hence, $f$ is \textbf{not} observed to be locally linear, (2) However, $|| f(x + \delta) - f(x)||_2$ is strongly correlated to $||f(\delta)||_2$, and our data-free optimization approach exploits this correlation between the two quantities. To summarize, our data-free optimization exploits the correlation between the quantities, $|| f(x + \delta) - f(x)||_2$ and $||f(\delta)||_2$, rather than the local-linearity of feature extractor $f$.

Finally, the relative change caused by our perturbations at the input to classification layer ($fc_8$ or softmax) can be clearly related to the fooling rates achieved for various perturbations. Table~\ref{tab:rel-shift-fooling} shows the relative shift in the feature activations $(\frac {\Vert l_i(x+\delta) - l_i(x) \Vert_2 } {\Vert l_i(x) \Vert_2})$ that are input to the classification layer and the corresponding fooling rates for various perturbations. Note that they are highly correlated, which explains why the proposed objective can fool the CNNs trained across multiple vision tasks.

%We depict fooling as a stability issue with the learned representations by CNNs. We attempt to learn the optimal perturbation in the input space that can cause maximal change in the output of the network. We achieve this via learning perturbations that can result in maximal change in the activations at all the intermediate layers of the architecture. {\color{blue} One may interpret that 'fooling' is achieved due to the locally linear nature of CNNswhy the proposed objective (eq.~\ref{eqn:data-free-objective}) works is that it due to CNNs being locally linear, i.e., $f(x + \delta) \approx f(x) + f(\delta)$. However, our experiments reveal that the feature extractor $f$ might not be locally linear. we try to observe the relation between the quantities $|| f(x + \delta) - f(x)||_2$ and $||f(\delta)||_2$, where $||.||_2$ represents the $L_2$-norm, and $f(.)$ represents the output of the last convolution layer of the network. Figure~\ref{fig:corrresnet} presents the comparison of these two quantities for ResNet-152 computed during the proposed optimization with no prior. The values computed are averaged over $5000$ random class-balanced images from Imagenet validation data. The perturbations are extracted before each rescale operation (projection into the $l_{\infinity}$ ball via division by $2$) during the optimization.

\begin{table}[t]
\centering
\caption{Relative Shift in Classification Layer's input vs. fooling rate for \textbf{VGG-16} for object recogntiion.
The relative shift has been evaluation on $1000$ random images from ILSVRC validation set, while the fooling rate is evaluated on the entire ILSVRC validation set.}
\label{tab:rel-shift-fooling}
\begin{tabular}{lcc}
\toprule
Perturbation &\pbox{20cm}{ Rel. Shift in input to \\ $fc_8$ (classification) layer}     & Fooling rate \\ \midrule 
Baseline & 0.0006 & 8.62 \\
No Prior & 0.867 & 45.47\\
Range Prior & 1.142 & 63.08\\
All data Prior & 3.169 & 77.77 \\\midrule
\end{tabular}
\end{table}
%\fi

\section{Conclusion}
\label{sec:conclu}

In this paper, we have proposed a novel data-free objective to craft image-agnostic (universal) adversarial perturbations (UAP). More importantly, we show that the proposed objective is generalizable not only across multiple CNN architectures but also across diverse computer vision tasks. We demonstrated that our seemingly simple objective of injecting maximal ``adversarial" energy into the learned representations (subject to the imperceptibility constraint) is effective to fool both the classification and regression models. Significant transfer performances achieved by our crafted perturbations can pose substantial threat to the deep learned systems in terms of black-box attacking. 

Further, we show that our objective can exploit minimal priors about the target data distribution to craft stronger perturbations. For example, providing simple information such as the mean and dynamic range of the images to the proposed objective would craft significantly stronger perturbations. Though the proposed objective is data-free in nature, it can craft stronger perturbations when data is utilized.

More importantly, we introduced the idea of generalizable objectives to craft image-agnostic perturbations. It is already established that the representations learned by deep models are susceptible. On top of it, the existence of generic objectives to fool ``any" learning based vision model independent of the underlying task can pose critical concerns about the model deployment. Therefore, it is an important research direction to be focused on in order to build reliable machine learning based systems.

%#####################################
% ######           Fig-11 Failure cases for Recognition  begins       #####################################
\begin{figure*}[ht]
\centering
\noindent\begin{minipage}{\textwidth}
  \centering
  \begin{minipage}{.16\textwidth}
  	\centering
    \includegraphics[width=\linewidth]{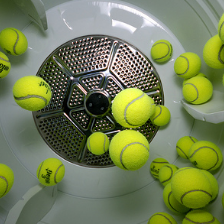}\
  \end{minipage}
   \begin{minipage}{.16\textwidth}
   	\centering
    \includegraphics[width=\linewidth]{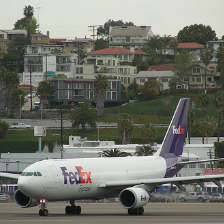}\
  \end{minipage}
  \begin{minipage}{.16\textwidth}
  	\centering
    \includegraphics[width=\linewidth]{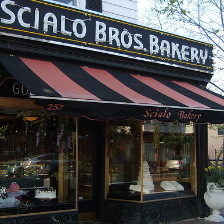}\
  \end{minipage}
  \begin{minipage}{.16\textwidth}
  	\centering
    \includegraphics[width=\linewidth]{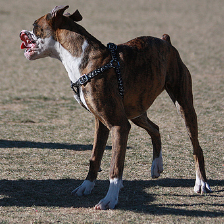}\
  \end{minipage}
  \begin{minipage}{.16\textwidth}
  	\centering
    \includegraphics[width=\linewidth]{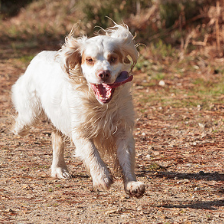}\
  \end{minipage}
  \begin{minipage}{.16\textwidth}
  	\centering
    \includegraphics[width=\linewidth]{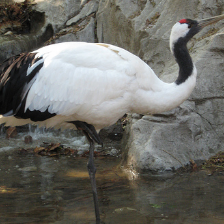}\
  \end{minipage}
  \vspace{0.002\textwidth}
\end{minipage}
\noindent\begin{minipage}{\textwidth}
  \centering
  \begin{minipage}{.16\textwidth}
  	\centering
    \includegraphics[width=\linewidth]{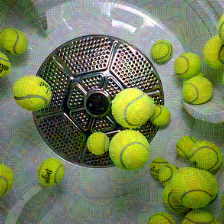}\
  \end{minipage}
   \begin{minipage}{.16\textwidth}
   	\centering
    \includegraphics[width=\linewidth]{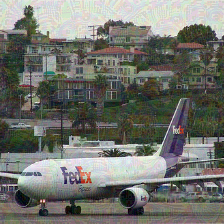}\
  \end{minipage}
  \begin{minipage}{.16\textwidth}
  	\centering
    \includegraphics[width=\linewidth]{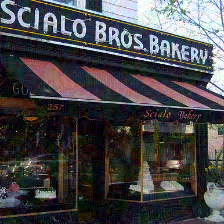}\
  \end{minipage}
  \begin{minipage}{.16\textwidth}
  	\centering
    \includegraphics[width=\linewidth]{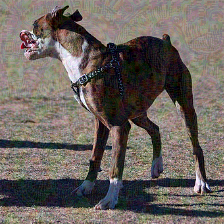}\
  \end{minipage}
  \begin{minipage}{.16\textwidth}
  	\centering
    \includegraphics[width=\linewidth]{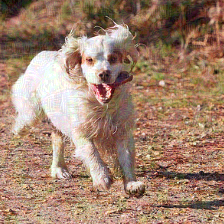}\
  \end{minipage}
  \begin{minipage}{.16\textwidth}
  	\centering
    \includegraphics[width=\linewidth]{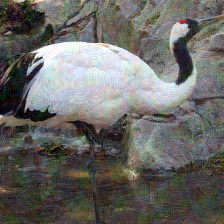}\
  \end{minipage}
  \vspace{0.002\textwidth}
\end{minipage}
\vspace{0.1cm}  
\caption{Sample failure case for the object recognition using VGG-16 model. Top row shows multiple clean images from ILSVRC validation set. Bottom row shows the adversarial images generated by adding the perturbation crafted utilizing the no data prior. Note that for all the shown images the perturbation fails to change the predicted label.}
\label{fig:recognition-failure-cases}
\end{figure*}
% ######           Fig-9  Failure cases for Recognition ends         #####################################
%#####################################
% ######           Fig-10 Failure cases for segmentation begins       #####################################
\begin{figure*}[ht]
\centering
\noindent\begin{minipage}{\textwidth}
  \centering
  \begin{minipage}{.18\textwidth}
  	\centering
    \includegraphics[width=\linewidth]{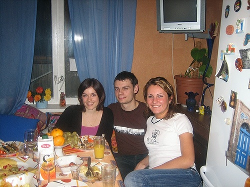}\
  \end{minipage}
   \begin{minipage}{.18\textwidth}
   	\centering
    \includegraphics[width=\linewidth]{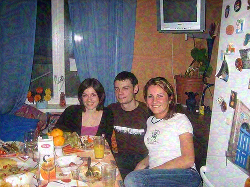}\
  \end{minipage}
  \begin{minipage}{.18\textwidth}
  	\centering
    \includegraphics[width=\linewidth]{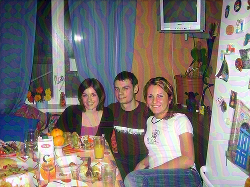}\
  \end{minipage}
  \begin{minipage}{.18\textwidth}
  	\centering
    \includegraphics[width=\linewidth]{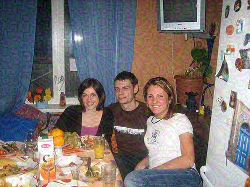}\
  \end{minipage}
  \begin{minipage}{.18\textwidth}
  	\centering
    \includegraphics[width=\linewidth]{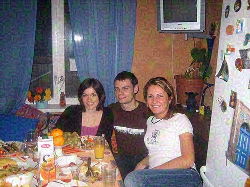}\
  \end{minipage}
  \vspace{0.002\textwidth}
\end{minipage}
\noindent\begin{minipage}{\textwidth}
  \centering
  \begin{minipage}{.18\textwidth}
  	\centering
    \includegraphics[width=\linewidth]{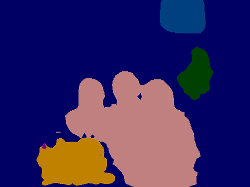}\
    Clean
  \end{minipage}
   \begin{minipage}{.18\textwidth}
   	\centering
    \includegraphics[width=\linewidth]{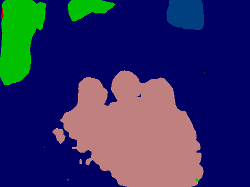}\
    No data
  \end{minipage}
  \begin{minipage}{.18\textwidth}
  	\centering
    \includegraphics[width=\linewidth]{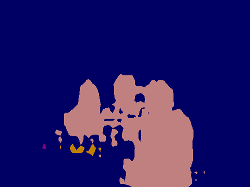}\
    Range prior
  \end{minipage}
  \begin{minipage}{.18\textwidth}
  	\centering
    \includegraphics[width=\linewidth]{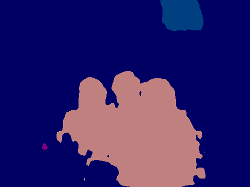}\
    Data w/ less BG
  \end{minipage}
  \begin{minipage}{.18\textwidth}
  	\centering
    \includegraphics[width=\linewidth]{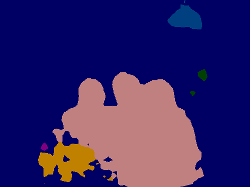}\
    All data
  \end{minipage}
\end{minipage}
\vspace{0.1cm}
\noindent\begin{minipage}{\textwidth}
\centering
\includegraphics[width=.93\linewidth]{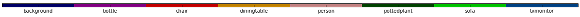}\
\end{minipage}
\caption{Sample failure case for the semantic segmentation using the FCN-8s-VGG model. Top row shows the clean and corresponding perturbed images for no prior to various priors. Bottom row shows the predicted segmentation maps. Note that the people segments are undisturbed by the addition of various perturbations.}
\label{fig:segmentation-failure-cases}
\end{figure*}
% ######           Fig-10  Failure cases for segmentation ends         #####################################
%#####################################
% ######           Fig-11 Failure cases for depth  begins       #####################################
\begin{figure*}[ht!]
\centering
\noindent\begin{minipage}{\textwidth}
  \centering
  \begin{minipage}{.23\textwidth}
  	\centering
    \includegraphics[width=\linewidth]{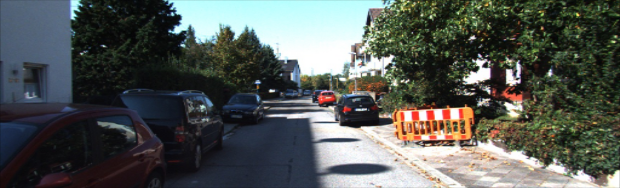}\
  \end{minipage}
   \begin{minipage}{.23\textwidth}
   	\centering
    \includegraphics[width=\linewidth]{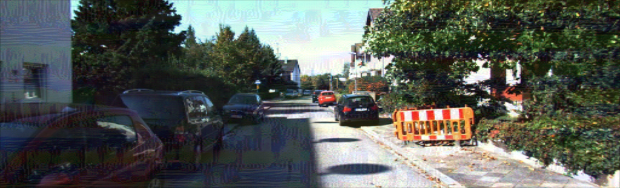}\
  \end{minipage}
  \begin{minipage}{.23\textwidth}
  	\centering
    \includegraphics[width=\linewidth]{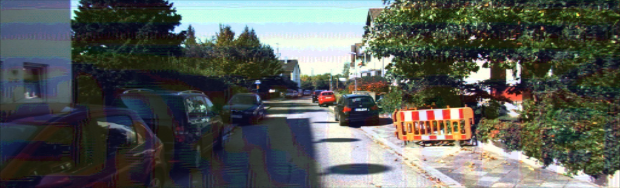}\
  \end{minipage}
  \begin{minipage}{.23\textwidth}
  	\centering
    \includegraphics[width=\linewidth]{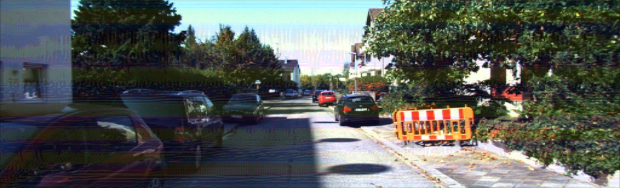}\
  \end{minipage}
  \vspace{0.002\textwidth}
\end{minipage}
\noindent\begin{minipage}{\textwidth}
  \centering
  \begin{minipage}{.23\textwidth}
  	\centering
    \includegraphics[width=\linewidth]{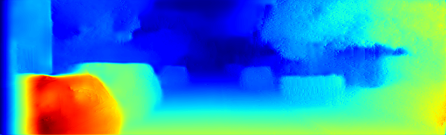}\
    Clean
  \end{minipage}
   \begin{minipage}{.23\textwidth}
   	\centering
    \includegraphics[width=\linewidth]{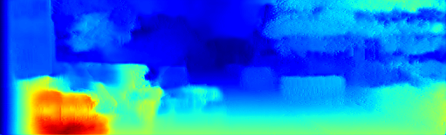}\
    No data
  \end{minipage}
  \begin{minipage}{.23\textwidth}
  	\centering
    \includegraphics[width=\linewidth]{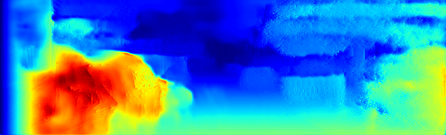}\
    Range prior
  \end{minipage}
  \begin{minipage}{.23\textwidth}
  	\centering
    \includegraphics[width=\linewidth]{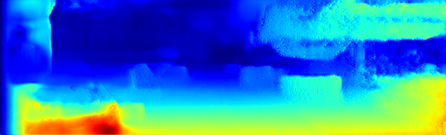}\
    Data prior
  \end{minipage}
  \vspace{0.002\textwidth}
\end{minipage}
\vspace{0.1cm}  
\caption{Sample failure case for the depth estimation using Monodepth-VGG model. Note that, top row shows clean image and the corresponding perturbed images with no data case and various prior cases. Bottom row shows the corresponding depth predictions.}
\label{fig:depth-failure-cases}
\end{figure*}
% ######           Fig-11  Failure cases for depth ends         #####################################

% if have a single appendix:
%\appendix[Proof of the Zonklar Equations]
% or
%\appendix  % for no appendix heading
% do not use \section anymore after \appendix, only \section*
% is possibly needed

% use appendices with more than one appendix
% then use \section to start each appendix
% you must declare a \section before using any
% \subsection or using \label (\appendices by itself
% starts a section numbered zero.)
%

%\appendices
%\section{Proof of the First Zonklar Equation}
%Appendix one text goes here.

% you can choose not to have a title for an appendix
% if you want by leaving the argument blank
%\section{}
%Appendix two text goes here.

% use section* for acknowledgment
%\ifCLASSOPTIONcompsoc
  % The Computer Society usually uses the plural form
%  \section*{Acknowledgments}
%\else
  % regular IEEE prefers the singular form
%  \section*{Acknowledgment}
%\fi

%The authors would like to thank...

% Can use something like this to put references on a page
% by themselves when using endfloat and the captionsoff option.
\ifCLASSOPTIONcaptionsoff
  \newpage
\fi
% trigger a \newpage just before the given reference
% number - used to balance the columns on the last page
% adjust value as needed - may need to be readjusted if
% the document is modified later
%\IEEEtriggeratref{8}
% The "triggered" command can be changed if desired:
%\IEEEtriggercmd{\enlargethispage{-5in}}

% references section

% can use a bibliography generated by BibTeX as a .bbl file
% BibTeX documentation can be easily obtained at:
% http://mirror.ctan.org/biblio/bibtex/contrib/doc/
% The IEEEtran BibTeX style support page is at:
% http://www.michaelshell.org/tex/ieeetran/bibtex/
\bibliographystyle{IEEEtran}
% argument is your BibTeX string definitions and bibliography database(s)
\bibliography{guap-bibliography.bib}
% if have a single appendix:
%\appendix[Proof of the Zonklar Equations]
% or
%\appendix  % for no appendix heading
% do not use \section anymore after \appendix, only \section*
% is possibly needed
% use appendices with more than one appendix
% then use \section to start each appendix
% you must declare a \section before using any
% \subsection or using \label (\appendices by itself
% starts a section numbered zero.)
\iffalse
\appendices
\section{Proof of the First Zonklar Equation}
Appendix one text goes here.
% you can choose not to have a title for an appendix
% if you want by leaving the argument blank
\section{}
Appendix two text goes here.
\fi
% Can use something like this to put references on a page
% by themselves when using endfloat and the captionsoff option.
\ifCLASSOPTIONcaptionsoff
  \newpage
\fi
\vspace{-0.2cm}
\begin{IEEEbiography}[{\includegraphics[width=0.8in,height=1.25in,clip,keepaspectratio]{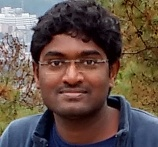}}]{Konda Reddy Mopuri} received an M.Tech degree  from  the  IIT Kharagpur, India in 2011. He is currently pursuing Ph.D. degree with the Department of Computational and Data Sciences, Indian Institute of Science (IISc), Bangalore, India, advised by Prof. R. Venkatesh Babu.  He  worked in Samsung India, Bangalore from  2011  to  2012. His research interests include computer vision, and machine learning with an emphasis on studying the deep learned visual representations.
\end{IEEEbiography}
\vspace{-0.2cm}
\begin{IEEEbiography}[{\includegraphics[width=0.8in,height=1.25in,clip,keepaspectratio]{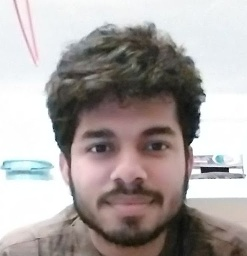}}]{Aditya Ganeshan} is a project assistant at Video Analytics Lab, Indian Institute of Science, Bangalore. He received his Integrated Master of Science in Applied Mathematics from Indian Institute of Technology, Roorkee. His research Interest includes machine learning, reinforcement learning and functional analysis. 
\end{IEEEbiography}
\vspace{-0.2cm}
\begin{IEEEbiography}[{\includegraphics[width=0.8in,height=1.25in,clip,keepaspectratio]{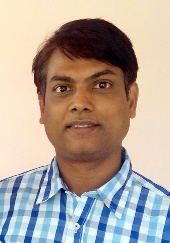}}]{R. Venkatesh Babu} received  the  Ph.D.  degree  from  the  Department  of Electrical  Engineering,  Indian  Institute  of  Science  (IISc),  Bangalore,  India. He held post-doctoral position with the Norwegian University of Science and Technology, Norway, and IRISA/INRIA, Rennes, France. He was a Research Fellow  with  Nanyang  Technological  University,  Singapore.  Currently,  he  is an  Associate  Professor  at  Dept.  of  Computational  and  Data  Sciences,  IISc. His  research  interests  span  signal  processing,  compression,  machine  vision, image/video  processing,  machine  learning,  and multimedia.
\end{IEEEbiography}

% You can push biographies down or up by placing
% a \vfill before or after them. The appropriate
% use of \vfill depends on what kind of text is
% on the last page and whether or not the columns
% are being equalized.

%\vfill

% Can be used to pull up biographies so that the bottom of the last one
% is flush with the other column.
%\enlargethispage{-5in}

% that's all folks
\end{document}